\documentclass[10pt, twocolumn, letterpaper]{article}

\usepackage[T1]{fontenc}
\usepackage{times}            % Times New Roman (neutral, professional)

\usepackage[
  top=20mm, bottom=25mm, left=16mm, right=16mm,
  columnsep=6mm
]{geometry}

\usepackage{amsmath, amssymb, amsfonts}
\usepackage{amsthm}
\newtheorem{definition}{Definition}

\usepackage{graphicx}
\graphicspath{{figures/}}
\usepackage[caption=false,font=small]{subfig}

\usepackage{booktabs}
\usepackage{makecell}
\usepackage{multirow}
\usepackage{array}

\usepackage{algorithm}
\usepackage{algorithmic}

\usepackage{textcomp}
\usepackage{stfloats}         % allow bottom-of-page full-width floats
\usepackage{verbatim}
\usepackage{cite}
\usepackage{bm}
\usepackage{xcolor}

\usepackage{url}
\usepackage[breaklinks,colorlinks,
  linkcolor=black, citecolor=black, urlcolor=blue]{hyperref}

\newcommand{\govA}{\textbf{A}\,}
\newcommand{\govP}{\textbf{P}\,}

\usepackage{fancyhdr}
\begin{document}

\title{S5-SHB Agent: Society 5.0 Enabled Multi-Model\\
Agentic Blockchain Framework for Smart Home}

\author{
  Janani Rangila$^{1}$, \
  Akila Siriweera$^{2,*}$, \
  Incheon Paik$^{2}$, \
  Keitaro Naruse$^{2}$, \\
  Isuru Jayananda$^{1}$, \
  Vishmika Devindi$^{1}$ \\[6pt]
  $^{1}$KD University, Colombo, Sri Lanka \qquad
  $^{2}$University of Aizu, Fukushima, Japan \\[4pt]
  \texttt{$^{*}$Corresponding: asiriwe@u-aizu.ac.jp}
}

\date{}  % suppress date under title

\twocolumn[
  \begin{@twocolumnfalse}
    \maketitle
    \begin{abstract}
%1. Importance of the problem
% The smart home is a key application domain within the Society 5.0 vision for a human-centered society. As smart home ecosystems expand with heterogeneous IoT protocols, diverse devices, and evolving threats, autonomous systems must manage comfort, security, energy, and safety for residents. Such autonomous decision-making requires a trust anchor, making blockchain a suitable foundation for transparent and accountable smart home governance.
% 
% The smart home is a key application domain within the Society 5.0 vision for a human-centered society. As smart home ecosystems expand with heterogeneous IoT and evolving threats, autonomous systems must manage comfort, security, energy, and safety for residents.
The smart home is a key application domain within the Society 5.0 vision for a human-centered society. 
As smart home ecosystems expand to include heterogeneous IoT devices and evolving threats, and diverse autonomous-systems requirements, such as security, privacy, comfort, and energy.
% for residents.
% 
Such decision-making requires a trust anchor, making blockchain a preferred foundation for transparent and accountable smart home governance.
%2. Clear problem description
% However, realizing this vision demands that blockchain-governed smart homes simultaneously address adaptive consensus, intelligent multi-agent coordination, and resident-controlled governance aligned with Society 5.0 principles.
% However, realizing this vision requires blockchain-governed smart homes to simultaneously address adaptive consensus, intelligent multi-agent coordination, and resident-controlled governance aligned with the principles of Society 5.0. 
%3. Issues/limitations with current solutions
% Existing blockchain-based smart home frameworks fall short on these requirements: they rely on fixed consensus protocols that cannot adapt to mixed transaction workloads, employ at most single-model AI without multi-agent coordination, and provide no tiered governance empowering residents to control automation behaviour.
% Existing rigid, smart contract-driven blockchain-based smart home frameworks fall short of these requirements: they rely on inflexible smart contracts, fixed consensus protocols that cannot adapt to dynamic situations, such as transaction workloads, employ at most a single-model AI without multi-agent coordination, and provide no tiered governance empowering residents to control automation behavior.
Existing frameworks rely on rigid smart contracts with fixed consensus protocols, employ mostly uncoordinated AI models, and offer no governance mechanism for residents to control automation.
% behavior.
%4. Our solution
% To address these limitations, this paper presents the Society 5.0 Agentic Blockchain-Governed Smart Home (S5-SHB-Agent). The framework coordinates ten specialized agents through interchangeable large language models (LLMs) for decision-making across safety, security, comfort, energy, privacy, and health domains. An adaptive PoW blockchain adjusts mining difficulty based on transaction volume and emergency conditions, with Ed25519 digital signatures and Merkle tree anchoring for tamper-evident auditability. A four-tier governance model enables residents to control automation through tiered preferences from routine adjustments to immutable safety thresholds.
% To address these limitations, this paper presents the Society 5.0-driven human-centered governance-enabled smart home blockchain agent (S5-SHB-Agent).
% The framework orchestrates ten specialized agents using interchangeable large language models to make decisions across the safety, security, comfort, energy, privacy, and health domains. An adaptive PoW blockchain adjusts the mining difficulty based on transaction volume and emergency conditions, with digital signatures and Merkle tree anchoring to ensure tamper-evident auditability. 
% A four-tier governance model enables residents to control automation through tiered preferences from routine adjustments to immutable safety thresholds.
To address these limitations, we present a Society 5.0-driven human-centered governance-enabled smart home blockchain agent (S5-SHB-Agent).
The framework orchestrates ten specialized agents—seven domain LLM agents covering safety, security, privacy, energy, climate, health, and maintenance, an ML-based anomaly detection agent,
% employing an Isolation Forest and Z-score ensemble, 
a natural language understanding agent, 
and an intelligent arbitration agent that resolves inter-agent conflicts.
% through a four-stage cascade of safety override, LLM reasoning, ML outcome scoring, and priority-based fallback—using interchangeable large language models across multiple providers. 
An adaptive proof-of-work blockchain adjusts mining difficulty based on transaction volume and emergency conditions, using Ed25519 digital signatures and Merkle tree anchoring for tamper-evident auditability. 
A four-tier governance model enables residents to control automation from routine adjustments to immutable safety thresholds.
%5. Evaluation / Achievement
% Evaluation across five dimensions demonstrates that adaptive consensus reduces emergency latency, the multi-agent architecture produces coherent cross-domain decisions, and the governance model enforces resident preferences while preserving safety invariants.
Evaluation confirms that governance correctly separates adjustable comfort priorities from immutable safety thresholds across all tested configurations; adaptive consensus commits emergency blocks in under 10 ms; the arbitration mechanism resolves all inter-agent conflicts with safety decisions consistently preserved; and across interchangeable LLM variants, the multi-agent system sustains 99–100\% decision acceptance with agent confidence consistently above 0.82. 
% (Source code available\footnote{Source code available at: https://github.com/AsiriweLab/S5-SHB-Agent})
    \end{abstract}
    \vspace{3mm}
    \vspace{3mm}
    \hrule
    \vspace{2mm}
    {\small
      \noindent
      Source code: \url{https://github.com/AsiriweLab/S5-SHB-Agent} (MIT License) \hfill
      Preprint --- not peer-reviewed
    }
    \vspace{6mm}
  \end{@twocolumnfalse}
]

% ============================================================
% BODY — \input individual section files
% ============================================================

% ============================================================
% Section 1: Introduction
% ============================================================
\section{Introduction}
\label{sec:intro}

The Japanese Cabinet Office's vision for Society 5.0 is a human-centered society that integrates cyber and physical systems to improve social well-being alongside economic progress \cite{CabinetOffice, siriweera2025autobda, 9371712}. 
The smart home is the primary residence where people interact with interconnected devices for security, energy management, health monitoring, and indoor climate control \cite{9612081, 10301586}. As these ecosystems expand, bridging heterogeneous device protocols and an ever-shifting threat landscape, the research community requires frameworks that govern such environments autonomously while remaining transparent and aligned with human values \cite{9371712, 9612081, 10301586, MUSAMIH2026131100}.

% Blockchain technology provides cryptographic non-repudiation, tamper-evident logging, and decentralized consensus for IoT device transactions \cite{alruwaili2024decentralized, erukala2025end, alruwaili2024blockchain}. Within the smart home domain, blockchain-based security has been addressed through deep learning-enhanced intrusion detection \cite{alruwaili2024decentralized, alruwaili2024blockchain} and consortium-based end-to-end secure communication \cite{erukala2025end}. Beyond the smart home, blockchain-based IoT frameworks address smart grid monitoring \cite{yang2019blockchain, kharbouch2025digital, mubarak2025fpga}, consumer electronics privacy \cite{kumar2026privacy, wang2026blockchain}, reinforcement learning-based intrusion detection \cite{yang2025qb}, healthcare IoT \cite{khan2025blockchain, kumar2026sbtm}, and general IoT concerns including DAG-based consensus \cite{de2026selecting}, scalable architectures \cite{xia2025scalable}, secure OTA firmware logging \cite{arshad2025secure}, transaction fee minimization \cite{jandaeng2024transaction}, privacy-preserving localization \cite{zhu2025privacy}, energy-efficient consensus \cite{jhariya2026energy}, quantum-assisted forensics \cite{khan2026quantum}, smart city communication \cite{guo2026integrated, fei2023sec}, and IoMT-based interoperability \cite{akram2025secure}.

\begin{figure}
\centering
\includegraphics[width=\columnwidth]{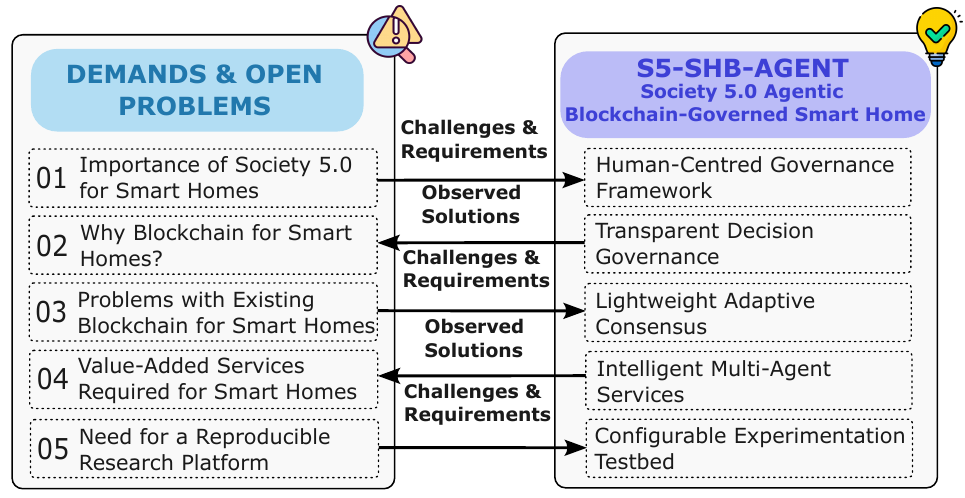}
\caption{Gap-driven design motivation for S5-SHB-Agent}
\label{fig:motivation}
\end{figure}

Blockchain technology has been widely applied to IoT security, offering 
% cryptographic non-repudiation, 
tamper-evident logging and decentralized consensus across domains ranging from smart home intrusion detection and healthcare to smart grids, smart cities, and general IoT infrastructure \cite{
10301586, 
% 11413910,
ZHANG2026130682, BUSHNAG2026130168,
% G1: 
alruwaili2024decentralized, erukala2025end, alruwaili2024blockchain,
% G2: 
akram2025secure,
% G3: 
yang2019blockchain, kharbouch2025digital}. 
Existing works demonstrate that blockchain can secure device transactions and audit logs, treat consensus as a fixed parameter, AI as an isolated model, and resident interaction as an access control problem, leaving the deeper challenge of human-centered, adaptive, multi-agent governance unaddressed to date \cite{yang2025qb, wang2026blockchain, kumar2026privacy, khan2026quantum}.

% Despite these efforts, a structured review of 21 recent blockchain-based IoT frameworks \cite{de2026selecting}-\cite{akram2025secure} reveals five research gaps that prevent existing solutions from fulfilling Society 5.0's human-centered vision: \textit{First, no framework addresses Society 5.0 human-centered socio-technical framing.} All 21 surveyed works treat blockchain as purely technical infrastructure without considering how residents should retain agency over automation decisions. Current frameworks offer at best basic access control \cite{erukala2025end, khan2025blockchain, akram2025secure} with no tiered governance for resident-controlled preferences.

Despite these efforts, a systematic review of recent blockchain-based IoT frameworks \cite{
% G1: 
alruwaili2024decentralized, erukala2025end, alruwaili2024blockchain,
% G2: 
akram2025secure,
% G3: 
yang2019blockchain, kharbouch2025digital, mubarak2025fpga,
% G4: 
wang2026blockchain, kumar2026privacy, yang2025qb, khan2026quantum,
% G5: 
khan2025blockchain, kumar2026sbtm,
% G6: 
de2026selecting, xia2025scalable, arshad2025secure, jandaeng2024transaction, zhu2025privacy, jhariya2026energy,
% G7: 
guo2026integrated, fei2023sec}
reveals five research gaps that prevent existing solutions from fulfilling Society 5.0's human-centered vision: 
First, no counterpart for Society 5.0's human-centered governance.
% All surveyed works treat blockchain as purely technical infrastructure without considering how residents should retain agency over automation decisions. 
Current frameworks offer, at best, basic access control \cite{erukala2025end, khan2025blockchain, akram2025secure} with no governance of resident-controlled preferences.

% \textit{Second, all 21 frameworks employ fixed consensus protocols incapable of runtime adaptation.} Whether using PoW \cite{alruwaili2024decentralized, alruwaili2024blockchain}, Proof-of-Stake, Proof-of-Authority, or DAG/Tangle variants \cite{de2026selecting}, every surveyed blockchain imposes static difficulty regardless of workload dynamics. This creates a mismatch where routine telemetry logging coexists with safety-critical emergency commands requiring rapid confirmation.

Second, all surveyed frameworks employ fixed consensus protocols incapable of runtime adaptation. 
Whether using Proof-of-Work (PoW) \cite{alruwaili2024decentralized, alruwaili2024blockchain}, Proof-of-Stake, Proof-of-Authority, or DAG/Tangle variants \cite{de2026selecting}, thus resulted in static difficulty regardless of workload dynamics. This creates a mismatch in which routine telemetry logging coexists with safety-critical emergency commands that require rapid confirmation.

% \textit{Third, no framework employs multi-agent systems orchestrated by multi-model large language models.} The most advanced AI integration is hybrid double Q-learning with Bi-LSTM \cite{yang2025qb}. Others employ individual deep learning models \cite{alruwaili2024decentralized, alruwaili2024blockchain, kumar2026sbtm} or metaheuristic optimization \cite{de2026selecting, jhariya2026energy}. None use LLM reasoning or multi-agent coordination across competing domains.

Third, LLM-based multi-agent orchestration is entirely absent from the surveyed works.
The rigid reliance on smart contracts, which are precompiled encoded logic with fixed rules that cannot reason across existing domain events. 
A smoke alarm and a power surge occurring simultaneously demand contextual judgment that no deterministic condition can encode. 
One of the AI integrations found is hybrid double Q-learning with Bi-LSTM \cite{yang2025qb}, while others employ individual deep learning models \cite{alruwaili2024blockchain, kharbouch2025digital, kumar2026sbtm} or metaheuristic optimization \cite{de2026selecting, jhariya2026energy}. 
LLM reasoning and cross-domain agent coordination across safety, security, energy, and health remain unexplored. 
Nevertheless, a single fixed model cannot simultaneously meet the sub-second response requirements of safety-critical agents and 
% the 
cost constraints of routine automation.
% ; tier-constrained multi-model routing is specifically designed to address
% this 
% heterogeneity.

% \textit{Fourth, no framework provides a tiered governance model for resident-controlled preferences.} Existing governance is limited to consortium membership \cite{erukala2025end}, encryption-gated access \cite{khan2025blockchain}, or interoperability-level permissions \cite{akram2025secure}. No framework implements hierarchical preference tiers spanning routine boundaries through immutable safety thresholds.

Fourth, conflict resolution and resident-controlled governance remain unaddressed across existing works. Existing governance is limited to consortium membership \cite{erukala2025end}, encryption-gated access \cite{khan2025blockchain}, or interoperability-level permissions \cite{akram2025secure}. 
Governance is uniformly flat; no tier structure separates resident-adjustable comfort settings from immutable safety thresholds.

% Fifth, evaluation environments remain uniformly isolated across all surveyed works, with no platform spanning simulation, real, and hybrid deployment modes. Only \cite{arshad2025secure} employs a real testbed and \cite{mubarak2025fpga} combines simulation with FPGA testing, but neither offers a unified platform with user-selectable mode switching. Fig.ref{fig:motivation} illustrates how these five demands map to solution components in S5-SHB-Agent.

% \textit{Fifth, no framework offers user-selectable multi-mode deployment spanning simulation, real, and hybrid environments.} The majority of surveyed works use simulation-only evaluation. Only \cite{arshad2025secure} employs a real testbed and \cite{mubarak2025fpga} combines simulation with FPGA testing, but neither offers a unified platform with user-selectable mode switching. Fig.ref{fig:motivation} illustrates how these five demands map to solution components in S5-SHB-Agent.

Fifth, no framework offers user-selectable multi-mode deployment spanning simulation, real, and hybrid environments. The majority of surveyed works use simulation-only evaluation. Only \cite{arshad2025secure} employs a real testbed and \cite{mubarak2025fpga} combines simulation with FPGA testing, but neither offers a unified platform with user-selectable mode switching.

\begin{table}[t]
\centering
\caption{Summary of notation.}
\label{tab:notations}
\begin{tabular}{p{0.18\columnwidth} p{0.72\columnwidth}}
\hline
\centering \textbf{Symbol} & \textbf{Definition} \\
\hline
\centering $\delta_t$       & Mining difficulty at block $t$ \\
\centering $\bar{v}_t$      & Avg.\ transaction count over sliding window of $w$ blocks \\
\centering $\mathcal{A}$    & Set of specialized agents $\{a_1, a_2, \ldots, a_n\}$ \\
\centering $\mathcal{T}(t)$ & Device telemetry vector at time $t$ \\
\centering $\pi(a_i)$       & Continuous priority score of agent $a_i$ \\
\centering $\sigma_i$       & Role-specific system prompt for agent $a_i$ \\
\centering $\tau(m_j)$      & Capability tier of model $m_j$ \\
\centering $\mathcal{R}$    & Tier-constrained model routing function \\
\centering $\mathcal{CR}$   & Four-level conflict resolution cascade \\
\centering $\mathcal{S}(a)$ & Composite ML score for agent $a$ \\
\centering $\mathcal{G}$    & Governance tier set $\{\Gamma_1,\Gamma_2,\Gamma_3,\Gamma_4\}$ \\
\centering $\phi$           & Governance authorization function \\
\centering $\mathcal{M}_r$  & Merkle root over off-chain records \\
\hline
\end{tabular}
\end{table}

%% ── Wide comparison table — spans both columns ────────────────────────────────
\begin{table*}[t]
\centering
\caption{Grouped comparison of blockchain-based IoT framework capabilities.}
\label{tab:grouped_comparison}
\renewcommand{\arraystretch}{1.3}
\setlength{\tabcolsep}{3pt}
\newcommand{\citeSmall}[1]{{\mbox{\scriptsize\cite{#1}}}}
\resizebox{\textwidth}{!}{%
\begin{tabular}{l|ccccc|ccc|ccccccc|cccc|cccc}
\toprule
\multicolumn{1}{c|}{\textbf{References}}
& \multicolumn{5}{c|}{\textbf{Application Domain}}
& \multicolumn{3}{c|}{\textbf{Blockchain}}
& \multicolumn{7}{c|}{\textbf{AI / ML Integration}}
& \multicolumn{4}{c|}{\textbf{Human-Center Govern}}
& \multicolumn{4}{c}{\textbf{Deployment Mode}} \\
& \rotatebox{75}{\makecell{SH}}
& \rotatebox{75}{\makecell{SG}}
& \rotatebox{75}{\makecell{HC}}
& \rotatebox{75}{\makecell{CE}}
& \rotatebox{75}{\makecell{SC/IoT}}
& \rotatebox{75}{\makecell{Fixed}}
& \rotatebox{75}{\makecell{Adapt}}
& \rotatebox{75}{\makecell{Cross}}
& \rotatebox{75}{\makecell{None}}
& \rotatebox{75}{\makecell{ML}}
& \rotatebox{75}{\makecell{Deep}}
& \rotatebox{75}{\makecell{RL}}
& \rotatebox{75}{\makecell{LLM}}
& \rotatebox{75}{\makecell{MultiA}}
& \rotatebox{75}{\makecell{MultiM}}
& \rotatebox{75}{\makecell{None}}
& \rotatebox{75}{\makecell{BAC}}
& \rotatebox{75}{\makecell{TierGov}}
& \rotatebox{75}{\makecell{S\,5.0}}
& \rotatebox{75}{\makecell{Sim}}
& \rotatebox{75}{\makecell{Real}}
& \rotatebox{75}{\makecell{Hybrid}}
& \rotatebox{75}{\makecell{Multi}} \\
\cline{1-24}
{G1: \cite{alruwaili2024decentralized,erukala2025end,alruwaili2024blockchain}}
  & $\checkmark$ & -- & -- & -- & --
  & $\checkmark$ & -- & --
  & \citeSmall{erukala2025end} & -- & \citeSmall{alruwaili2024decentralized,alruwaili2024blockchain} & -- & -- & -- & --
  & \citeSmall{alruwaili2024decentralized,alruwaili2024blockchain} & \citeSmall{erukala2025end} & -- & --
  & $\checkmark$ & -- & -- & -- \\
{G2: \cite{akram2025secure}}
  & $\checkmark$ & -- & $\checkmark$ & -- & --
  & $\checkmark$ & -- & $\checkmark$
  & -- & -- & $\checkmark$ & -- & -- & -- & --
  & -- & $\checkmark$ & -- & --
  & $\checkmark$ & -- & -- & -- \\
{G3: \cite{yang2019blockchain,kharbouch2025digital,mubarak2025fpga}}
  & -- & $\checkmark$ & -- & -- & --
  & $\checkmark$ & -- & --
  & \citeSmall{yang2019blockchain} & \citeSmall{mubarak2025fpga} & \citeSmall{kharbouch2025digital} & -- & -- & -- & --
  & $\checkmark$ & -- & -- & --
  & $\checkmark$ & $\checkmark$ & -- & -- \\
{G4: \cite{wang2026blockchain,kumar2026privacy,yang2025qb,khan2026quantum}}
  & -- & -- & -- & $\checkmark$ & --
  & $\checkmark$ & -- & --
  & -- & $\checkmark$ & $\checkmark$ & $\checkmark$ & -- & -- & --
  & $\checkmark$ & -- & -- & --
  & $\checkmark$ & -- & -- & -- \\
{G5: \cite{khan2025blockchain,kumar2026sbtm}}
  & -- & -- & $\checkmark$ & -- & --
  & $\checkmark$ & -- & --
  & -- & -- & $\checkmark$ & -- & -- & -- & --
  & \citeSmall{kumar2026sbtm} & \citeSmall{khan2025blockchain} & -- & --
  & $\checkmark$ & -- & -- & -- \\
{G6: \cite{de2026selecting,xia2025scalable,arshad2025secure,jandaeng2024transaction,zhu2025privacy,jhariya2026energy}}
  & -- & -- & -- & -- & $\checkmark$
  & $\checkmark$ & -- & --
  & \citeSmall{xia2025scalable,arshad2025secure} & \citeSmall{de2026selecting,jandaeng2024transaction,zhu2025privacy,jhariya2026energy} & -- & -- & -- & -- & --
  & $\checkmark$ & -- & -- & --
  & $\checkmark$ & $\checkmark$ & -- & -- \\
{G7: \cite{guo2026integrated,fei2023sec}}
  & -- & -- & -- & -- & $\checkmark$
  & $\checkmark$ & -- & --
  & $\checkmark$ & -- & -- & -- & -- & -- & --
  & $\checkmark$ & -- & -- & --
  & $\checkmark$ & -- & -- & -- \\

\textbf{Ours}
  & $\checkmark$ & -- & -- & -- & --
  & -- & $\checkmark$ & $\checkmark$
  & -- & -- & $\checkmark$ & -- & $\checkmark$ & $\checkmark$ & $\checkmark$
  & -- & -- & $\checkmark$ & $\checkmark$
  & $\checkmark$ & $\checkmark$ & $\checkmark$ & $\checkmark$ \\
\bottomrule
\end{tabular}%
}
\end{table*}

% To address these gaps, we present the Society 5.0 Agentic Blockchain-Governed Smart Home (S5-SHB-Agent). The framework adopts a three-layer architecture: a Control Plane managing external interfaces and governance policies; an Agent Intelligence layer coordinating ten specialized AI agents (seven domain agents and three extension agents) through interchangeable LLMs with a multi-model router and four-level conflict resolution cascade; and a Device \& Data layer encapsulating IoT device abstraction via Model Context Protocol (MCP) alongside blockchain-anchored trust infrastructure. An adaptive PoW blockchain adjusts mining difficulty based on transaction volume, combined with firmware-level emergency bypass for immediate safety response, Ed25519 digital signatures, and Merkle tree anchoring for tamper-evident auditability. A four-tier governance model aligned with Society 5.0 enables residents to control automation through tiered preferences without programming expertise. To the best of our knowledge, S5-SHB-Agent is the first framework that simultaneously addresses Society 5.0 framing, adaptive consensus, multi-agent multi-model LLM orchestration, tiered human-centered governance, and multi-mode deployment for blockchain-enabled smart home IoT.

To address these gaps, we present the Society 5.0 Agentic Blockchain-Governed Smart Home (S5-SHB-Agent), built on a layered architecture comprising a \textit{control panel, agentic intelligence, blockchain, devices,} and \textit{external services.}
% , built on three architectural layers. 
% The Control Plane manages external interfaces and enforces governance policies. The Agent Intelligence layer coordinates 10 specialized AI agents (seven domains, three extensions) via a multi-model LLM router with tier-constrained provider assignment and a four-level conflict-resolution. The Device \& Data layer abstracts IoT hardware through the Model Context Protocol alongside blockchain-anchored trust infrastructure. An adaptive Proof-of-Work blockchain scales mining difficulty with transaction volume. Firmware-level emergency bypass keeps safety commands instantaneous; Ed25519 digital signatures and Merkle anchoring produce unforgeable audit records. A four-tier governance model enables residents to shape automation, from daily comfort preferences to immutable safety thresholds, without requiring configuration expertise.
% 
Fig.\ref{fig:motivation} illustrates how five demands map to solution components in S5-SHB-Agent.

Our key contributions are:
\begin{itemize}
    \item 
    % A four-tier human-centered governance model aligned with Society 5.0, where residents control automation through safe adjustments, impactful trade-offs, advanced overrides, and immutable safety thresholds via natural language interfaces.
    A four-tier human-centered governance model aligned with Society 5.0; residents control automation through routine adjustments, advanced overrides, and immutable safety thresholds via natural-language interfaces.

    \item 
    % An adaptive PoW blockchain that adjusts mining difficulty based on transaction volume, complemented by firmware-level emergency bypass for immediate safety-critical response, with Ed25519 digital signatures and Merkle tree anchoring for tamper-evident auditability.
    An adaptive PoW blockchain that adjusts mining difficulty based on transaction volume, complemented by firmware-level emergency bypass for immediate safety-critical response, with Ed25519 digital signatures and Merkle tree anchoring for tamper-evident auditability.
    
    \item 
    % A multi-agent orchestration architecture coordinating ten specialized AI agents across four priority tiers through a multi-model LLM router supporting four providers with tier-constrained model assignment.
    A multi-agent orchestration architecture coordinating ten specialized AI agents across four priority tiers through a multi-model LLM router supporting four providers with tier-constrained model assignment.
    
    \item 
    % A four-level conflict resolution cascade resolving inter-agent disputes through safety override, LLM-based contextual arbitration, ML-based historical scoring, and priority-based fallback.
    A four-level conflict resolution resolving inter-agent disputes through safety override, LLM-based contextual arbitration, ML-based historical scoring, and priority-based fallback.

    \item 
    % A multi-mode deployment framework supporting simulation, real, and hybrid environments with a unified orchestration layer and tamper-evident Merkle anchoring.
    A multi-mode deployment framework supporting simulation, real, and hybrid environments with a unified orchestration layer and tamper-evident Merkle anchoring.

\end{itemize}

% To the best of our knowledge, S5-SHB-Agent is the first smart-contract-free blockchained agentic framework that simultaneously addresses Society 5.0 framing, adaptive consensus, multi-agent multi-model LLM orchestration, tiered human-centered governance, and multi-mode deployment (simulation, real, and hybrid) for smart home IoT.

To the best of our knowledge, S5-SHB-Agent is the first smart-contract-free agentic-blockchain framework governed by Society 5.0 principles.
% framing, adaptive consensus, multi-agent multi-model LLM orchestration, tiered human-centered governance, and multi-mode deployment (simulation, real, and hybrid) for smart home IoT.
% 
% Table I outlines frequently used acronyms. The remainder of the paper is structured as follows. Section \ref{sec_2} reviews related work. Section \ref{sec_3} outlines preliminaries including the motivation scenario and problem formulation. Section \ref{sec_4} presents the proposed S5-SHB-Agent framework. Section \ref{sec_5} discusses system demonstration. Section \ref{sec_6} presents evaluation results. Section \ref{sec_7} concludes with future work. 
% 
Table I summarizes the notation table. The remainder of the paper is structured as follows. 
Section \ref{sec_2} reviews related works. 
% Section \ref{sec_3} outlines preliminaries, including the motivation scenario and problem formulation. 
Section \ref{sec_3} outlines the motivation scenario and problem formulation. 
S5-SHB-Agent framework is presented in Section \ref{sec_4}.
% Section \ref{sec_4} presents the proposed S5-SHB-Agent framework. 
Section \ref{sec_5} discusses system evaluation. 
% Section \ref{sec_5} presents conclusion. 
Finally, conclusions with future work in Section \ref{sec_5}.
% ============================================================
% Section 2: Related Works
% ============================================================
\section{Related Works}
\label{sec_2}

\noindent This section surveys twenty-one recent blockchain-based IoT publications from 2025 
\cite{
% G1: 
alruwaili2024decentralized, erukala2025end, alruwaili2024blockchain,
% G2: 
akram2025secure,
% G3: 
yang2019blockchain, kharbouch2025digital, mubarak2025fpga,
% G4: 
wang2026blockchain, kumar2026privacy, yang2025qb, khan2026quantum,
% G5: 
khan2025blockchain, kumar2026sbtm,
% G6: 
de2026selecting, xia2025scalable, arshad2025secure, jandaeng2024transaction, zhu2025privacy, jhariya2026energy,
% G7: 
guo2026integrated, fei2023sec} 
spanning smart homes, smart grids, healthcare, consumer electronics, and general IoT security. We conducted a structured survey aligned with the research objective: \textit{investigating the holistic, inclusive, and intelligent features that accommodate Society 5.0's human-centered, agentic, and trust-anchored research paradigm for blockchain-enabled smart-home IoT}.  The resulting taxonomy captures three dimensions: \textit{holistic} (application domain coverage), \textit{inclusive} (blockchain architecture, AI/ML integration depth, and human-centered governance), and \textit{agile} (deployment mode flexibility). Table II presents the grouped comparison across these dimensions.

We identified seven groups (G1-G7) based on distinctive capability patterns. {G1: \cite{alruwaili2024decentralized, erukala2025end, alruwaili2024blockchain}} addresses smart home security using fixed consensus blockchains with traditional ML and deep learning, limited to simulation-only deployment. {G2: \cite{akram2025secure}} uniquely bridges smart home and healthcare domains through the only cross-chain interoperable architecture in the survey, with tiered access control but simulation-only evaluation. {G3: \cite{yang2019blockchain, kharbouch2025digital, mubarak2025fpga}} targets smart grid energy management combining traditional ML, metaheuristics, and deep learning; notably, \cite{mubarak2025fpga} provides one of only two real-testbed validations across all surveyed works. {G4: \cite{wang2026blockchain, kumar2026privacy, yang2025qb, khan2026quantum}} exhibits the most sophisticated single-framework AI, uniquely combining deep learning with reinforcement learning (hybrid Double Q-Learning and Bi-LSTM in \cite{yang2025qb}) and quantum-assisted techniques in \cite{khan2026quantum}, though confined to fixed consensus and simulation-only deployment.

{G5: \cite{khan2025blockchain, kumar2026sbtm}} applies blockchain to healthcare IoT with deep learning and encryption-based access control. {G6: \cite{de2026selecting, xia2025scalable, arshad2025secure, jandaeng2024transaction, zhu2025privacy, jhariya2026energy}}, the largest cluster, spans general IoT optimization and security; \cite{arshad2025secure} provides the second real-testbed validation alongside simulation. {G7: \cite{guo2026integrated, fei2023sec}} addresses smart city security with lightweight fixed-consensus blockchains and traditional ML. Despite progressive domain-specific advances, all seven groups share critical limitations: fixed consensus protocols, single-model AI at best, basic or absent governance, and predominantly simulation-only evaluation.

Five systematic research gaps emerge across all 21 works: (1) no Society 5.0 human-centered socio-technical framing, (2) no tiered governance enabling resident-controlled preferences, (3) no adaptive consensus capable of runtime difficulty adjustment, (4) no multi-agent LLM orchestration for intelligent conflict resolution, and (5) no user-selectable multi-mode deployment spanning simulation, real, and hybrid environments. Our proposed S5-SHB-Agent addresses all five gaps simultaneously: adaptive PoW consensus adjusting difficulty based on transaction volume, 10 specialized AI agents with multi-model LLM routing across four providers (Google Gemini, Anthropic Claude, OpenAI GPT, local Ollama), four-tier Society 5.0 governance with immutable safety invariants, and unified multi-mode deployment with tamper-evident Merkle anchoring capabilities absent from every surveyed framework.
% \begin{figure}
% \centering
% \includegraphics[width=1.0\columnwidth]{figures/Ref Ar.pdf}
% \caption{Reference architecture of S5-SHB Agent showing the three-layer design with Control Plane, Agent Intelligence, and Device \& Data layers.}
% \label{fig:ref_arch}
% \end{figure}

% =============================================================================

% ============================================================
% Section 3: Preliminary
% ============================================================
\section{Preliminary}
\label{sec_3}

This section formalises the smart-home governance scenario and states the problem addressed by S5-SHB-Agent.

\subsection{Smart-Home Governance Scenario}

A smart home $\mathcal{H}$ comprises a set of IoT devices $\mathcal{D}=\{d_1,\dots,d_n\}$ deployed across rooms, each producing periodic telemetry. A multi-agent layer $\mathcal{A}=\{a_1,\dots,a_k\}$ monitors telemetry and issues commands, while a blockchain $\mathcal{B}$ records every agent transaction with Ed25519 digital signatures and adaptive Proof-of-Work consensus.

The system supports three deployment modes \emph{simulation} (behavioural modelling via S5-HES-Agent), \emph{real} (protocol adapters: MQTT with QoS-1 and optional TLS, HTTP with bearer/basic/API-key authentication), and \emph{hybrid} (both simultaneously) unified through a Model Context Protocol (MCP) gateway that exposes nine tools as the single interface between agents and devices.

\subsection{Problem Definition}

\begin{definition}[Adaptive Proof-of-Work Consensus]\label{def:adaptive-pow}
Let $V_w$ denote the average transaction volume over a sliding window of $w=3$ blocks:
\begin{equation}\label{eq:avg_volume}
  V_w = \frac{1}{w}\sum_{i=1}^{w} |\mathit{tx}_i|
\end{equation}
The difficulty $\delta$ adjusts stepwise within $[\delta_{\min}=1,\;\delta_{\max}=4]$ relative to a base $\delta_b=2$, with volume thresholds $v_{\ell}=3$ and $v_{h}=10$:
\begin{equation}\label{eq:adaptive_consensus}
  \delta \leftarrow
  \begin{cases}
    \max(\delta_{\min},\;\delta - 1), & V_w > v_{h}\\[4pt]
    \min(\delta_{\max},\;\delta + 1), & V_w < v_{\ell}\\[4pt]
    \delta - \operatorname{sgn}(\delta - \delta_b), & \text{otherwise}
  \end{cases}
\end{equation}
\end{definition}

\begin{definition}[Agent Decision Function]\label{def:agent-decision}
Each agent $a_i$ has a static priority $\pi_i\in(0,1]$ and a processing function $f_i\in\{\mathrm{LLM},\mathrm{ML}\}$. Ten agents span seven domain roles and three cross-domain specialised roles:

\smallskip
\noindent\textbf{Domain agents} (LLM-based): Safety ($\pi{=}1.0$), Health ($\pi{=}0.9$), Security ($\pi{=}0.8$), Privacy ($\pi{=}0.7$), Energy ($\pi{=}0.6$), Climate ($\pi{=}0.5$), Maintenance ($\pi{=}0.4$).

\noindent\textbf{Cross-domain specialised}: NLU ($\pi{=}0.85$, LLM), Anomaly Detection ($\pi{=}0.88$, ML ensemble: Isolation Forest + LOF + Z-score + optional Autoencoder), Arbitration ($\pi{=}0.95$, LLM).

\smallskip
\noindent Given telemetry $T_t$ and governance parameters $G$, each agent produces a decision:
\begin{equation}\label{eq:agent_decision}
  a_i(T_t, G) \;\xrightarrow{f_i}\; \langle \mathit{action},\;\mathit{params},\;\mathit{confidence} \rangle
\end{equation}
Every decision is signed with Ed25519 and submitted as a blockchain transaction. The model router selects from eight models across four providers (Google Gemini, Anthropic Claude, OpenAI GPT, Ollama) in two tiers (flash/pro) with four presets (balanced, max\_privacy, budget, best\_quality):
\begin{equation}\label{eq:model_router}
  \mathit{model}(a_i) = \mathit{Router}(\mathit{preset},\;\mathit{tier}(a_i))
\end{equation}
\end{definition}

\begin{definition}[Four-Stage Conflict Resolution]\label{def:conflict}
When multiple agents target the same device, a four-stage arbitration cascade resolves conflicts:
\begin{equation}\label{eq:conflict_resolution}
  \mathit{resolve}(C) =
  \begin{cases}
    a_{\text{safety}}, & \text{safety override (immune)}\\
    a_{\text{LLM}}, & \text{LLM arbitration}\\
    \arg\max_{a} S(a), & \text{ML scoring}\\
    \arg\max_{a} \pi_a, & \text{priority fallback}
  \end{cases}
\end{equation}
where $S(a) = 0.6\,s_{\mathrm{ML}} + 0.4\,\pi_a$, with $s_{\mathrm{ML}}$ the ML-predicted acceptance score. The safety agent is never overridden at any stage.
\end{definition}

\begin{definition}[Four-Tier Governance]
\label{def:governance}
Resident preferences are partitioned into four tiers with monotonically increasing restriction. Let $\mathcal{P}$ denote all governance parameters and $\alpha(p)$ the authorisation level required to modify parameter $p$:

\begin{equation}\label{eq:governance}
  \alpha(p) \in
  \begin{cases}
    \text{SAFE (8 keys)}, & \text{freely adjustable}\\
    \text{IMPACTFUL (7 keys)}, & \text{requires confirmation}\\
    \text{ADVANCED (3 keys)}, & \text{expert override}\\
    \text{LOCKED (9 keys)}, & \text{immutable invariants}
  \end{cases}
\end{equation}
Twelve typed validation rules enforce range and choice constraints. Governance is implemented off-chain via a \texttt{GovernanceContract} Python class no on-chain smart contracts are used.
\end{definition}

\begin{definition}[Deployment Invariance]\label{def:deployment}
The orchestration loop maintains identical agent processing regardless of deployment mode $m$:
\begin{equation}\label{eq:deployment}
  \forall\; m \in \{\text{sim},\,\text{real},\,\text{hybrid}\}:\quad
  \mathcal{A}(T_t^{(m)}) = \mathcal{A}(T_t)
\end{equation}
The MCP gateway abstracts device access so that telemetry collection (every 10\,s) and agent cycles (every 20\,s with parallel LLM calls via \texttt{asyncio.gather}) are mode-independent. Firmware-level fallback rules in the Device Layer handle emergencies (smoke, gas, flood) autonomously, bypassing both the Agent Intelligence and Blockchain layers.
\end{definition}

% =============================================================================
% ============================================================
% Section 4: Proposed Method
% ============================================================
\section{Proposed Method: S5-SHB-Agent}
\label{sec_4}

% \noindent This section presents the S5-HSB-Agent designed to address the six challenges identified in Section \ref{sec_3}. Section \ref{sec_4_1} presents the reference architecture (RA). Section \ref{sec_4_2} describes the system architecture (SA). Sections \ref{sec_4_3} through \ref{sec_4_8} detail the solutions addressing each problem definition.

\noindent This section describes S5-SHB-Agent addressing the six challenges identified in Section \ref{sec_3}. Section \ref{sec_4_1} describes the system architecture, and Sections \ref{sec_4_2}--\ref{sec_4_7} detail each solution.

\subsection{System Architecture}
\label{sec_4_1}

% Fig. \ref{fig:sys_arch} instantiates the RA into deployable components. The Control Plane comprises a FastAPI backend (port 8001) with five WebSocket streams (telemetry, blockchain, agents, governance, scenarios) and a Vue.js/TypeScript frontend. The Agent Intelligence layer implements 7 Domain LLM Agents, NLU Agent, Arbitration Agent, and Anomaly Detection with the MCP gateway mediating device access through nine tools. The Device layer contains HESDevice (a universal class handling 118+ device types), an Emergency Scanner for firmware-level smoke and gas detection, and the DeviceLayer Manager. The Blockchain layer implements Adaptive PoW, Agent Registry, Conflict Detector, Ed25519 signing, off-chain store, and Merkle anchoring.

\begin{figure}
\centering
\includegraphics[width=1.0\columnwidth]{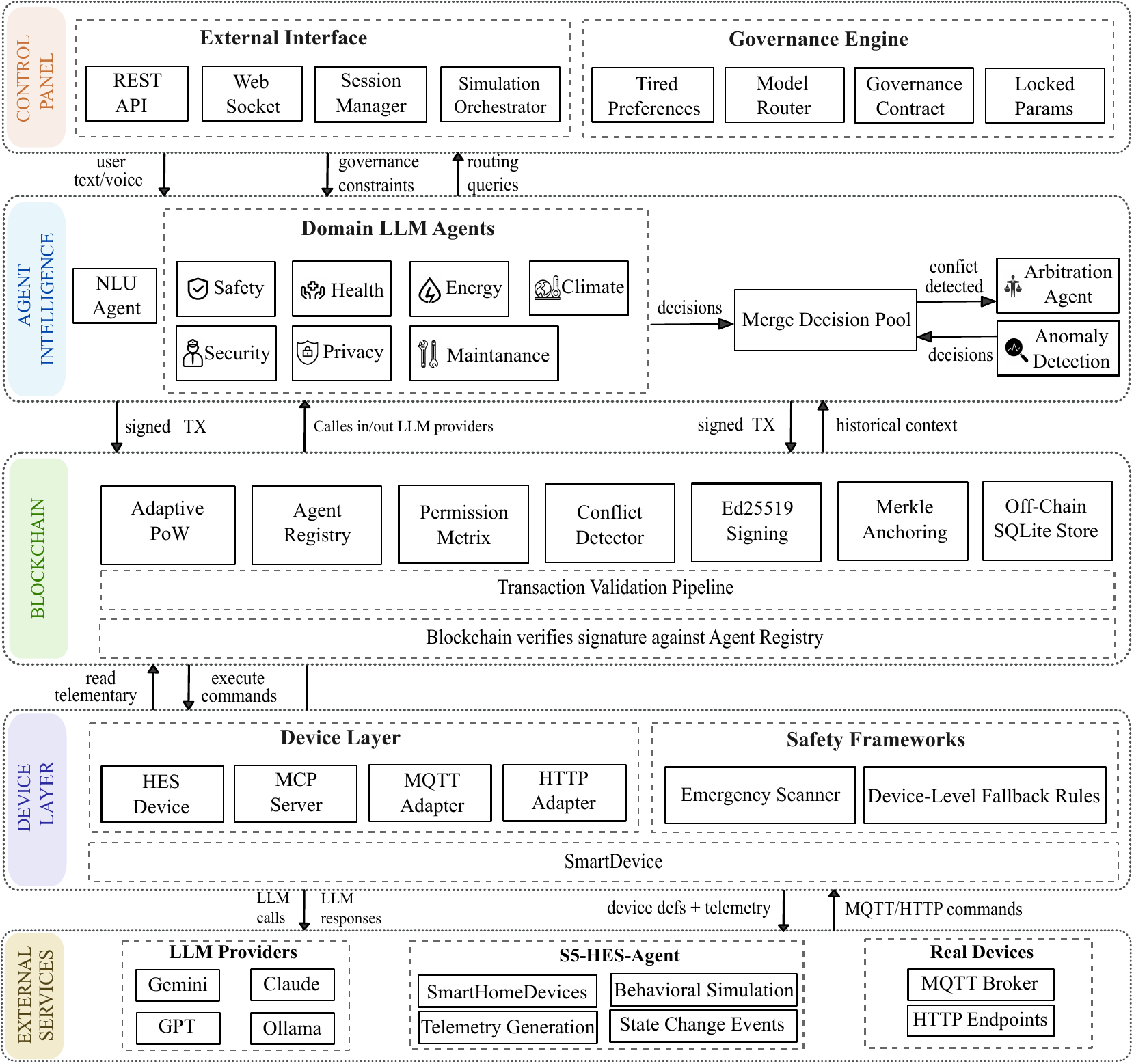}
\caption{System architecture of S5-SHB Agent showing the four-layer implementation with Control Plane, Agent Intelligence, Device \& Data, and External layers.}
\label{fig:sys_arch}
\end{figure}

Fig. \ref{fig:sys_arch} presents the system architecture in terms of deployable components. 
The Control Plane has a FastAPI backend running on port 8001, with 5 WebSocket streams, and a Vue.js/TypeScript frontend. The Agent Intelligence layer has 7 Domain LLM Agents, NLU, Arbitration, and Anomaly Detection, with the MCP gateway mediating access to devices using 9 tools. 
The Device layer has HESDevice \cite{siriweera2026s5hesagentsociety50driven}, a universal class supporting 118+ devices, Emergency Scanner, a firmware-based smoke and gas detector, and the DeviceLayer Manager. 
The Blockchain layer has Adaptive PoW, Agent Registry, Conflict Detector, Ed25519 signing, off-chain storage, and Merkle anchoring.

% An External layer makes explicit the dependencies: LLM Providers (Google Gemini, Anthropic Claude, OpenAI GPT, Ollama), S5-HES-Agent (port 8000) for behavioural simulation data in simulation mode, and real device infrastructure (MQTT brokers, HTTP endpoints) in real and hybrid modes. The Simulation Orchestrator coordinates the operational loop telemetry every 10\,s, agent cycles every 20\,s with parallel LLM calls via \text{asyncio.gather} identically across all deployment modes. Figs. \ref{fig:simulation_view}--\ref{fig:agent_monitor} present the dashboard interfaces demonstrating transparent oversight.

\begin{figure}
\centering
\includegraphics[width=\columnwidth]{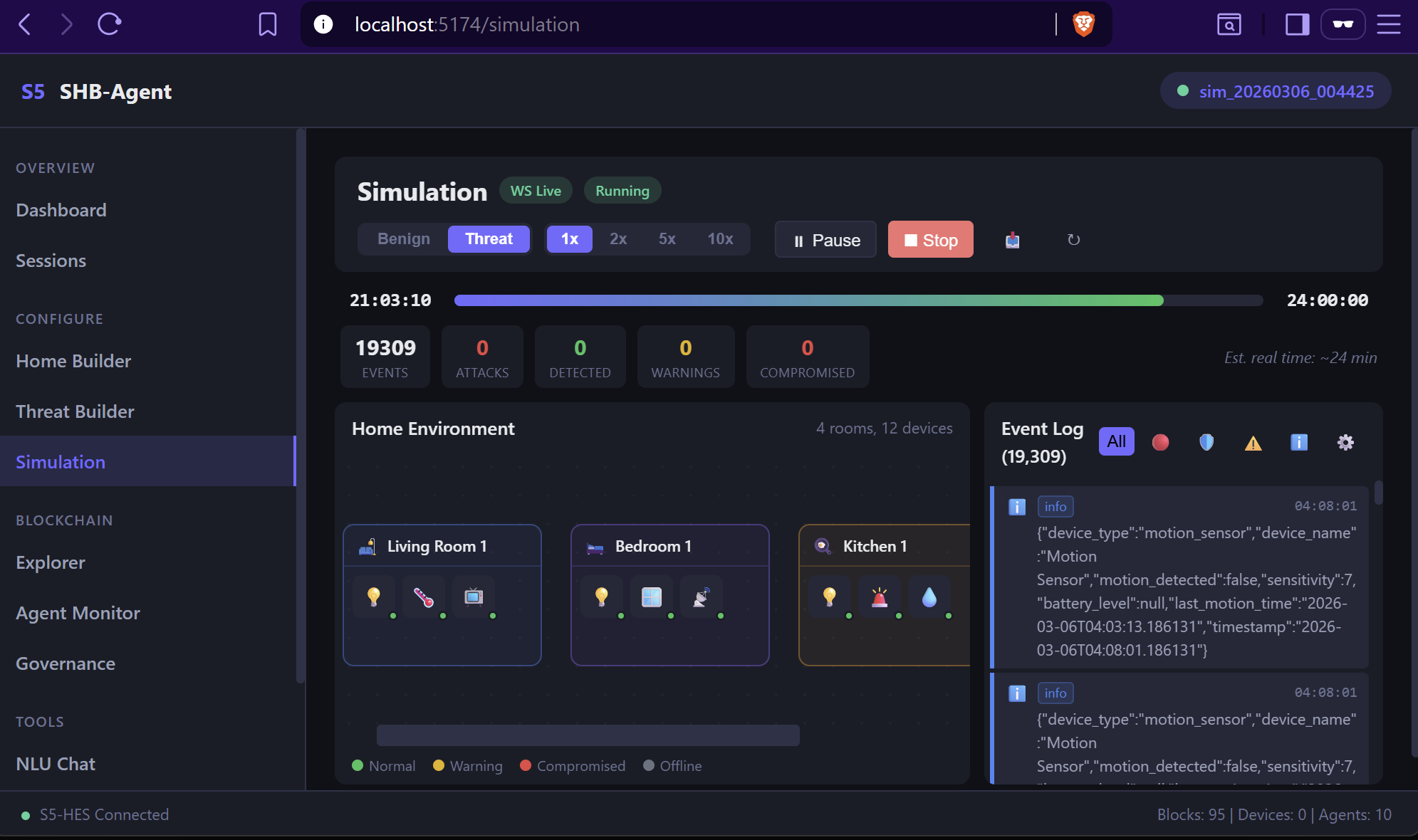}
\caption{Simulation View interface showing real-time threat scenario execution with home environment visualization and live event log streaming.}
\label{fig:simulation_view}
\end{figure}

\begin{figure}
\centering
\includegraphics[width=\columnwidth]{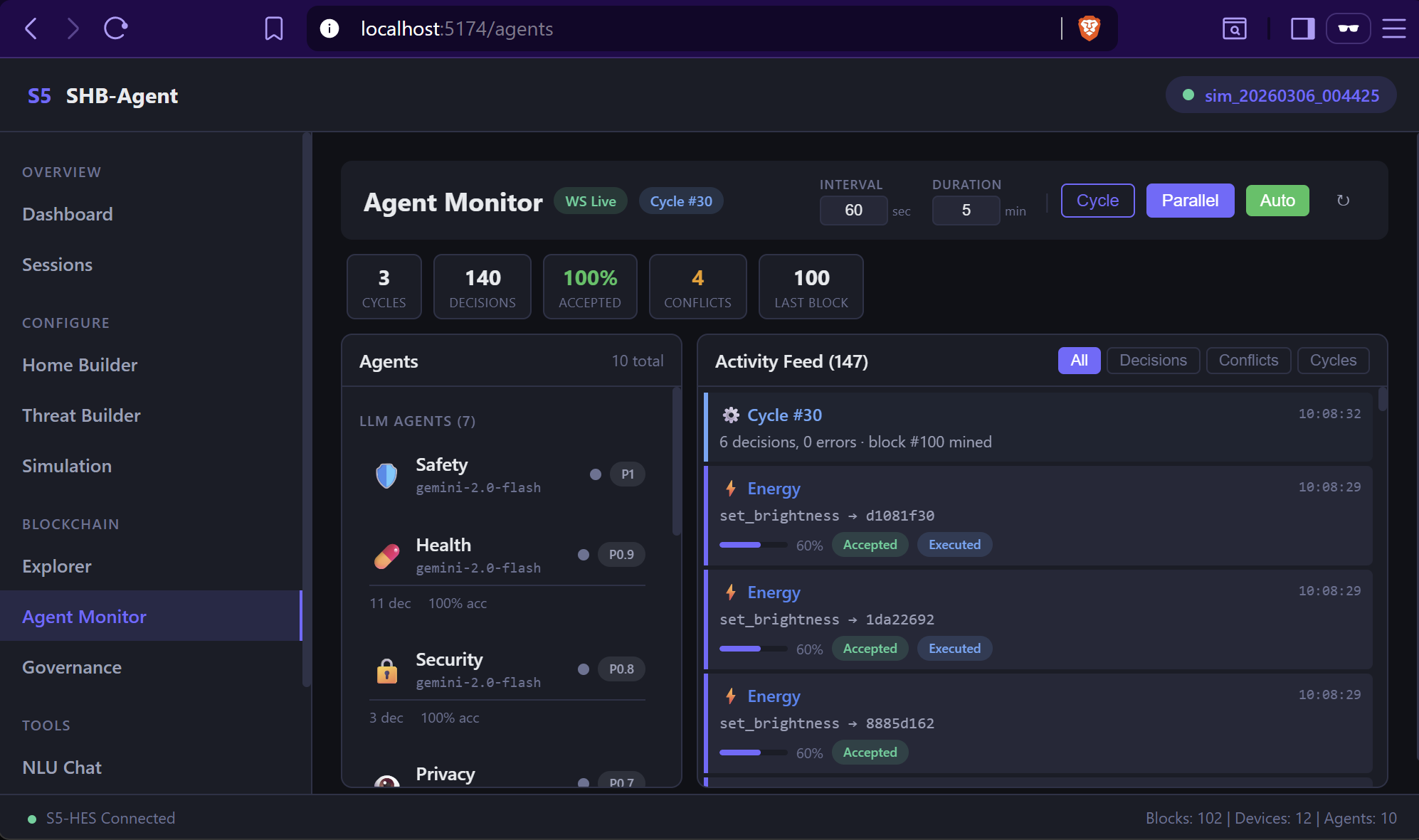}
\caption{Agent Monitor interface showing multi-agent reasoning cycles with per-agent decisions, confidence scores, and real-time activity feed.}
\label{fig:agent_monitor}
\end{figure}

An External layer makes explicit the dependencies: LLM Providers (Google Gemini, Anthropic Claude, OpenAI GPT, Ollama), S5-HES-Agent (port 8000) for behavioral simulation data in simulation mode, and real device infrastructure (MQTT brokers, HTTP endpoints) in real and hybrid modes. 
The Simulation Orchestrator manages the operational loop telemetry every 10\,s, agent cycles every 20\,s with parallel LLM calls through \text{asyncio.gather} identically across all deployment modes. 
Figs. \ref{fig:simulation_view}--\ref{fig:agent_monitor} show the dashboard interfaces to illustrate transparent oversight.

\subsection{Adaptive Consensus Blockchain (Solution to problem Definition 1)}
\label{sec_4_2}

% The blockchain maintains a lightweight chain where each block $B_t$ contains agent transactions, a SHA-256 hash linking to the previous block, a nonce satisfying the current difficulty, and a timestamp. The adaptive difficulty implements Equation (\ref{eq:adaptive_consensus}) through a sliding window estimator.

The blockchain has a lightweight chain where each block $B_t$ has the transactions made by the agents, a SHA-256 hash referencing the previous block, a nonce satisfying the current difficulty, and a timestamp. 
Adaptive difficulty makes use of Equation (\ref{eq:adaptive_consensus}) through a sliding window estimator.

Let $W = \{B_{t-w+1}, \ldots, B_t\}$ denote the $w$ most recent blocks (default $w{=}3$). The volume estimator computes:
\begin{equation}
\bar{V}(t) = \frac{1}{|W|} \sum_{B_i \in W} |B_i|
\label{eq:volume}
\end{equation}
where $|B_i|$ is the transaction count. Difficulty adjusts stepwise, clamped within $[\delta_{\min}{=}1,\;\delta_{\max}{=}4]$:
\begin{equation}
\delta(t\!+\!1) = \text{clamp}\!\Bigg(
\begin{cases}
\delta(t) - 1, & \text{if } \bar{V}(t) \geq v_{\text{high}} \\
\delta(t) + 1, & \text{if } \bar{V}(t) \leq v_{\text{low}}, \\
& \delta_{\min},\;\delta_{\max} \\
\delta(t) \mp 1 \to \delta_{\text{base}}, & \text{otherwise}
\end{cases}
\Bigg)
\label{eq:difficulty_adjust}
\end{equation}
where $v_{\text{high}}{=}10$, $v_{\text{low}}{=}3$, and $\delta_{\text{base}}{=}2$. High-volume periods, including emergency bursts from the Emergency Scanner (smoke ${\geq}\,0.3$, CO ${\geq}\,50$\,ppm, CO$_2$ ${\geq}\,5000$\,ppm, NG ${\geq}\,1000$\,ppm), organically lower difficulty for faster mining. The Emergency Scanner operates at the Device layer, ensuring safety response occurs \textit{before} AI reasoning.

% Every transaction carries an Ed25519 signature: $\text{sig}(tx_i) = \text{Sign}_{sk_{a_j}}(H(tx_i))$. A four-gate validation pipeline registration check, signature verification, permission check, and conflict detection validate each transaction. All agents hold wildcard device access at the permission level; role-based scoping is enforced through domain-specific LLM system prompts, ensuring no permission constraint blocks safety-critical cross-domain responses.

Each transaction has an Ed25519 signature: $\text{sig}(tx_i) = \text{Sign}_{sk_{a_j}}(H(tx_i))$. 
A four-gate pipeline for permission validation, signature verification, permission checking, and conflict detection is used for transaction validation. 
All agents have wildcard device access at the permission level, and role-based constraints are enforced by domain-specific LLM system prompts, which ensures that no permission constraint prevents safety-critical cross-domain access.

\subsection{Multi-Agent Decision Orchestration (Solution to problem Definition 2)}
\label{sec_4_3}

% Table \ref{tab:agent_specs} presents the complete agent specifications. Each Domain LLM Agent constructs a prompt from its role-specific system prompt $\sigma_i$, live device telemetry, and governance constraints:

The complete specification for the agents is provided in Table \ref{tab:agent_specs}. 
Each Domain LLM Agent generates a prompt based on its role-specific system prompt $\sigma_i$, device telemetry, and governance constraints: 
\begin{equation}
a_i(\mathcal{T}(t)) = \text{LLM}_i\big(\sigma_i \;\|\; \mathcal{T}(t) \;\|\; \mathcal{G}_{\text{constraints}}\big) \rightarrow (cmd_i, conf_i)
\label{eq:agent_reasoning}
\end{equation}
% where $\|$ denotes prompt concatenation and $conf_i \in [0,1]$ is the confidence score. All seven domain agents execute in parallel via \text{asyncio.gather} every 20\,s, independently analysing the same telemetry snapshot. Each command is SHA-256 hashed, Ed25519 signed, and submitted to the blockchain.
where $\|$ represents prompt concatenation and $conf_i \in [0,1]$ represents the confidence score. 
All seven domain agents are concurrently executed via \text{asyncio.gather} every 20\,s, where the same telemetry snapshot is independently processed. 
The command is SHA-256 hashed and Ed25519 signed before it is sent to the blockchain.

% The Anomaly Agent uses a four-model ML/DL ensemble (Isolation Forest, LOF, Z-score, optional autoencoder), flagging anomalies when two or more detectors agree or when the Isolation Forest score exceeds a critical threshold. Corrective actions are determined through device-specific mappings (7 device types plus a generic fallback). These corrective \text{AgentDecision} objects merge into the decision pool alongside domain agent decisions, entering the arbitration cascade only when conflicts arise. The NLU Agent parses natural-language commands, dynamically building a device catalogue from live telemetry and producing independent blockchain-signed decisions.

Anomaly Agent employs a four-model ML/DL ensemble (Isolation Forest, LOF, Z-score, and optional autoencoder), identifying anomalies by consensus of two or more models or an Isolation Forest score exceeding a critical threshold. 
Corrective actions are determined by device-specific mappings (7 device types and a generic fallback). 
The corrective \text{AgentDecision}s contribute to the pool of decisions, entering the cascade of arbitration after any conflict is encountered. 
The NLU Agent interprets natural language commands, creating a device catalogue from dynamic telemetry and providing independent decisions signed by blockchain.

% Device-level fallback rules adjusting temperature when readings exceed 30\textdegree{}C or fall below 16\textdegree{}C, and the Emergency Scanner provide continued safety operation during a complete LLM outage.

The device-level fallback rules, which adjust the temperature if it exceeds 30\textdegree{}C or drops below 16\textdegree{}C, and the Emergency Scanner ensure the continued safe operation of the system in the event of a complete LLM outage. 

\subsection{Multi-Model LLM Routing (Solution to problem Definition 3)}
\label{sec_4_4}

% The Model Router supports four providers (Google Gemini, Anthropic Claude, OpenAI GPT, Ollama) with eight models classified into tiers $\tau \in \{\textit{pro}, \textit{flash}\}$ and privacy levels $\rho \in \{\textit{cloud}, \textit{local}\}$. The routing function enforces tier constraints:
The Model Router has four providers, Google Gemini, Anthropic Claude, OpenAI GPT, and Ollama, with eight models organized into tiers $\tau \in \{\textit{pro}, \textit{flash}\}$ and privacy levels $\rho \in \{\textit{cloud}, \textit{local}\}$. The routing function enforces tier constraints:

\begin{equation}
\mathcal{R}(a_i) = \operatorname*{select}_{m_j \in \mathcal{M}} m_j, \quad \text{s.t. } \tau_j \geq \tau_{\min}(a_i)
\label{eq:routing}
\end{equation}
% where $\tau_{\min}$ maps agents to minimum model requirements: the Safety Agent requires ``pro'' tier (enforced as an immutable Tier 4 governance parameter); all other LLM agents accept ``flash'' or above; the Anomaly Agent uses ML models, bypassing tier constraints.
where $\tau_{\min}$ maps agents to minimum model requirements: the Safety Agent requires ``pro'' tier or better, enforced as an immutable Tier 4 governance parameter; all other LLM Agents require ``flash'' tier or better; the Anomaly Agent uses ML models, circumventing tier restrictions.

Per-agent cost tracking accumulates token usage:
\begin{equation}
\mathcal{C}_{\text{total}}(a_i) = \sum_{t=1}^{T} \mathcal{C}_{\text{token}}(m_{j(t)}) \cdot \big(|prompt_t| + |response_t|\big)
\label{eq:cost}
\end{equation}

% enabling budget monitoring through Tier 3 governance preferences. Four presets \textit{balanced} (pro for safety, flash for others), \textit{max\_privacy} (all local Ollama), \textit{budget} (cheapest with pro for safety), \textit{best\_quality} (pro everywhere) support profile switching through governance changes. When providers become unavailable, device-level fallback rules activate.
enabling budget monitoring via Tier 3 governance preferences. The four presets \textit{balanced} (pro for safety, flash for others), \textit{max\_privacy} (all local ollama), \textit{budget} (cheapest with pro for safety), and \textit{best\_quality} (pro everywhere) facilitate profile changes via governance changes. 
When providers become unavailable, device-level fallback rules are activated.

\subsection{Intelligent Conflict Resolution (Solution to problem Definition 4)}
\label{sec_4_5}

% When agents submit competing commands targeting the same device, the Arbitration Agent implements the four-stage cascade from Equation (\ref{eq:conflict_resolution}).
% \textbf{L1: Safety Override.} If either command originates from the Safety Agent ($\pi{=}1.0$), it wins unconditionally:
When different commands are submitted by agents on the same device, the Arbitration Agent follows a cascade of four stages as defined by Equation (\ref{eq:conflict_resolution}). \textbf{L1: Safety Override.} In the first level of arbitration, if either of the commands is sent by the Safety Agent ($\pi={=}1.0$), it wins unconditionally:

\begin{table}
\centering
\caption{Agent Specification with Priority}
\label{tab:agent_specs}
\begin{tabular}{p{0.19\columnwidth} p{0.22\columnwidth} p{0.20\columnwidth} p{0.23\columnwidth}}
\hline
\textbf{Agent $A_i$} & \textbf{Responsibility} & \textbf{Priority $\pi$} & \textbf{Output Domain} \\
\hline
Safety Agent & Fire, gas, smoke\par response & Safety Critical (1.0) & Emergency commands \\
Arbitration Agent & Conflict resolution & Safety Critical (0.95) & Resolution decisions \\
Health Agent & Occupant wellness monitoring & High (0.9) & Health alerts \\
Anomaly Agent & ML-based anomaly detection & High (0.88) & Anomaly alerts \\
NLU Agent & Natural language parsing & High (0.85) & Structured commands \\
Security Agent & Intrusion detection & Medium (0.8) & Lock/alarm actions \\
Privacy Agent & Camera management & Medium (0.7) & Privacy controls \\
Energy Agent & Power optimization & Standard (0.6) & Load scheduling \\
Climate Agent & Temperature, humidity & Standard (0.5) & HVAC commands \\
Maintenance Agent & Device health monitoring & Standard (0.4) & Maintenance alerts \\
\hline
\end{tabular}
\end{table}

\begin{equation} \mathcal{CR}_{L1}(cmd_i, cmd_j) = cmd_s, \quad \text{if } a_s = \text{safety-agent-001}
\label{eq:safety_override}
\end{equation}
% This is an immutable invariant architecturally enforced, not merely policy. % \textbf{L2: LLM Arbitration.} For non-safety conflicts, the Arbitration Agent invokes an LLM with full context:
This invariant cannot be violated, as it is architecturally enforced, not just policy. 
\textbf{L2: LLM Arbitration.} For non-safety conflicts, the Arbitration Agent will call an LLM with full context: 
\begin{equation}
cmd_{\text{winner}} = \mathcal{L}\big(cmd_i, cmd_j, \mathcal{T}(t), \mathcal{G}_{\text{prefs}}, \pi\big)
\label{eq:llm_arbitration}
\end{equation}

\textbf{L3: ML Scoring.} If LLM arbitration fails, historical performance scoring activates:
\begin{equation}
\mathcal{S}(a_i) = 0.6 \cdot \underbrace{r_{\text{accept}}(a_i) \cdot (1 - 0.5 \cdot r_{\text{conflict}}(a_i))}_{\text{ML historical score}} + 0.4 \cdot \pi(a_i)
\label{eq:ml_scoring}
\end{equation}

\textbf{L4: Priority Fallback.} As the deterministic backstop:
\begin{equation}
cmd_{\text{winner}} = cmd_{\operatorname{arg\,max}_{a_i} \pi(a_i)}
\label{eq:priority_fallback}
\end{equation}
Every conflict record, commands, resolution level, and winner rationale are logged to off-chain storage and anchored on-chain for auditing.

\subsection{Human-Centred Tiered Governance (Solution to problem Definition 5)}
\label{sec_4_6}

The Governance Contract manages the hierarchy $\mathcal{G} = \{G_1, G_2, G_3, G_4\}$ with monotonically increasing restriction (Table \ref{tab:governance_tiers}). The authorisation function:

\begin{figure}
\centering
\includegraphics[width=\columnwidth]{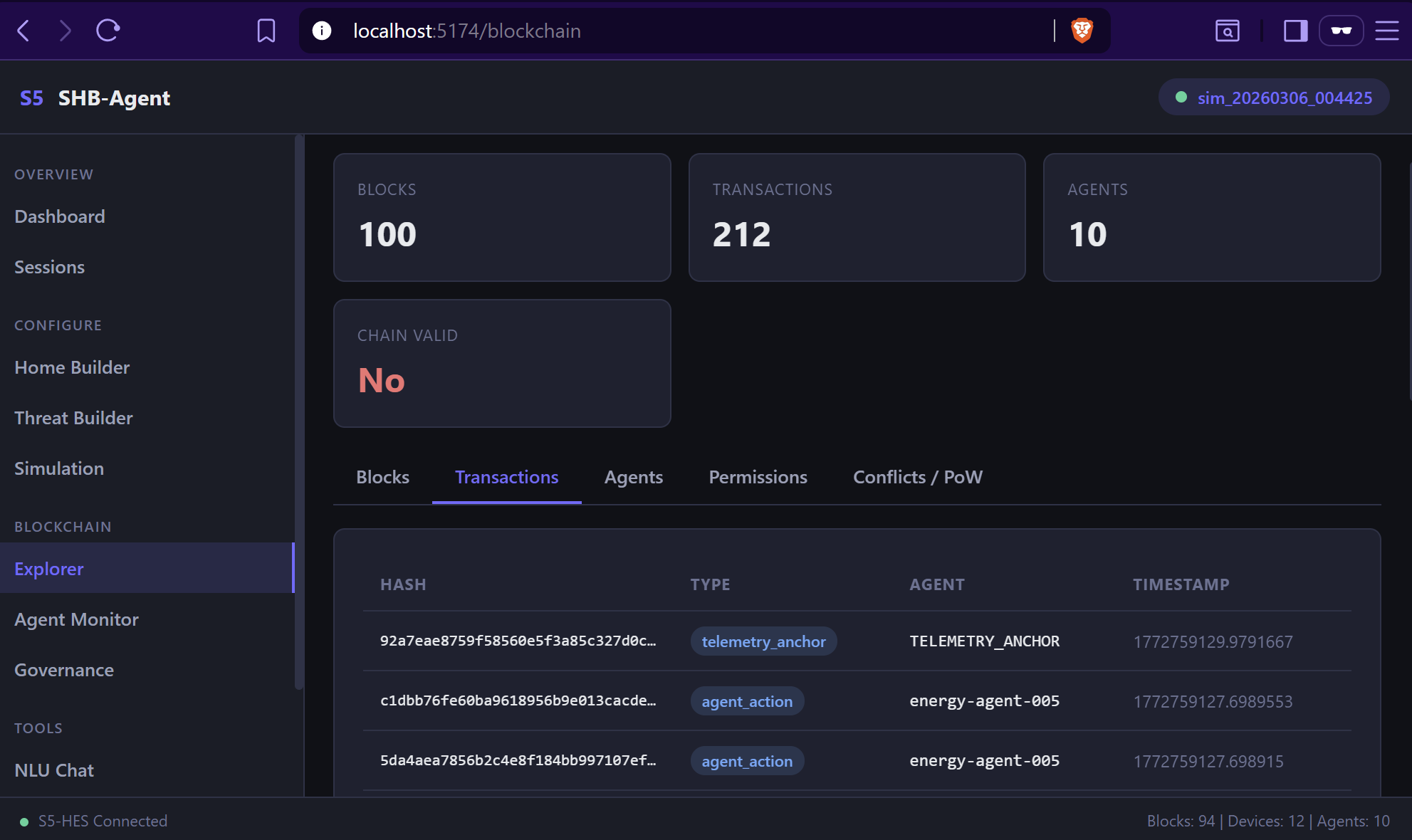}
\caption{Blockchain Explorer interface displaying the immutable transaction ledger with Ed25519-signed transactions, registered agents, and chain validity verification.}
\label{fig:blockchain_explorer}
\end{figure}

\begin{equation}
\phi(\mathcal{P}_u, G_k) =
\begin{cases}
\textit{allow}, & \text{if } k \leq 3 \text{ and } \text{valid}(v, R_k) \\
\textit{deny}, & \text{if } k = 4 \quad \forall u
\end{cases}
\label{eq:gov_impl}
\end{equation}

% where $\text{valid}(v, R_k)$ checks against 12 typed validation rules. Tiers 1--3 (8 SAFE, 7 IMPACTFUL, 3 ADVANCED keys) are resident-accessible with range/choice validation. Tier 4 (9 LOCKED keys) is computationally immutable the Governance Contract rejects any modification regardless of requester identity. Governance is implemented off-chain via a Python class; no on-chain smart contracts are used.
where $\text{valid}(v, R_k)$ verifies against 12 typed validation rules. Tiers 1--3 (8 SAFE, 7 IMPACTFUL, 3 ADVANCED keys) are resident-accessible with range/choice validation. Tier 4 (9 LOCKED keys) is computationally immutable with the Governance Contract refusing any change request regardless of the identity of the requester. 
The Governance mechanism is off-chain with a Python class, with no smart contracts used.

% When preferences change, updated constraints propagate to all agents via $\mathcal{G}_{\text{constraints}}$ in Equation (\ref{eq:agent_reasoning}). Tier 2 trade-off sliders dynamically influence conflict resolution outcomes. The NLU Agent enables natural-language governance ``I prefer comfort over energy savings'' translates to a Tier 2 parameter update with all changes validated and logged on-chain.
When preferences are updated, new constraints are propagated through all agents via $\mathcal{G}_{\text{constraints}}$ in Equation (\ref{eq:agent_reasoning}). Tier 2 trade-off sliders are dynamic and influence conflict resolution. 
The NLU Agent allows for natural language-based governance. For example, ``I prefer comfort over energy savings'' is a Tier 2 parameter update, where all changes are validated and logged on-chain.

\subsection{Solution to Definition 6: Multi-Mode Deployment}
\label{sec_4_7}

% The framework supports three modes $\mathcal{O} = \{\textit{sim}, \textit{real},\\
% \textit{hyb}\}$ through a unified interface. In simulation mode, the MCP tool interface provides device access. In real mode, \text{MQTTDeviceAdapter} (QoS-1, optional TLS) and \text{HTTPDeviceAdapter} (bearer/basic/API-key auth) extend the same \text{SmartDevice} base class, ensuring agents and blockchain are completely unaware of deployment mode. An \text{AdapterRegistry} maps protocols to adapter classes for extensibility.
This framework supports three modes $\mathcal{O} = \{\textit{sim}, \textit{real},\\
\textit{hyb}\}$ via a common interface. 
For the simulation mode, the MCP tool interface is used. For the real mode, the \text{MQTTDeviceAdapter} (with QoS-1, optional TLS) and the \text{HTTPDeviceAdapter} (with bearer, basic, and API-key auth) share the same base class, namely \text{SmartDevice}. 
This ensures agents and the blockchain have no idea about the mode.
The \textit{Simulation Orchestrator} follows a deterministic cycle:
% A deterministic cycle followed by \textit{Simulation Orchestrator}:
% The Simulation Orchestrator implements a deterministic cycle:
\begin{equation}
\begin{aligned}
\text{Cycle}(t) : \mathcal{T}(t) \xrightarrow{\text{store}} \mathcal{D}_{\text{off}} \xrightarrow{\text{reason}} \{cmd_i\} \xrightarrow{\text{resolve}} cmd^* \\\xrightarrow{\text{mine}} B_t
\label{eq:cycle}
\end{aligned}
\end{equation}
% where telemetry is collected every 10\,s, stored in the off-chain database $\mathcal{D}_{\text{off}}$ (13 tables), processed by agents every 20\,s, resolved through the cascade, and mined into block $B_t$.
where telemetry is collected every 10\,s, stored in the off-chain database $\mathcal{D}_{\text{off}}$ (13 tables), processed by agents every 20\,s, 
% resolved by the cascade, and then mined into block $B_t$.
resolved and mined into block $B_t$.
Merkle trees connect off-chain storage with on-chain integrity:
% Merkle trees bridge off-chain storage with on-chain integrity:
\begin{equation}
\mathcal{M}_r = \text{MerkleRoot}\big(\{H(rec_1), H(rec_2), \ldots, H(rec_n)\}\big)
\label{eq:merkle}
\end{equation}
% where $H(\cdot)$ is SHA-256. The Merkle root is committed on-chain, enabling tamper detection against any off-chain modification. The complete deployment function:
where $H(\cdot)$ is SHA-256. 
This Merkle commitment is tamper-detectable against any off-chain modification. 
The entire function for deployment:
\begin{equation}
\mathcal{S}_{\text{deploy}}(\omega, \mathcal{D}_{\text{off}}) = (\mathcal{B}_{\text{chain}}, \mathcal{M}_r), \quad \omega \in \mathcal{O}
\label{eq:deploy_final}
\end{equation}
% produces blockchain-anchored records with Merkle commitments regardless of deployment mode, addressing the flexibility gap identified across all 21 surveyed works.
produces blockchain-anchored records with a Merkle commitment, regardless of the chosen mode.
% , thus addressing the flexibility gap identified in all 21 works reviewed.

\begin{table}
\centering
\caption{Four-Tiered Governance Preference Hierarchy}
\label{tab:governance_tiers}
\begin{tabular}{p{0.14\columnwidth} p{0.16\columnwidth} p{0.30\columnwidth} p{0.22\columnwidth}}
\hline
\textbf{Tier} & \textbf{Scope} & \textbf{Parameters (Count)} & \textbf{Mutability} \\
\hline
$G_1$ \par(Safe) & Routine adjustments & Temperature, brightness, quiet hours, voice, alerts (8 keys) & Resident-modifiable \\
$G_2$ \par(Impactful) & Trade-off settings & Comfort-vs-energy, security-vs-privacy, automation level (7 keys) & Resident-modifiable with validation \\
$G_3$ \par(Advanced) & System \par overrides & Per-agent device overrides, API budget caps, allowed providers (3 keys) & Resident-modifiable with validation \\
$G_4$ (Locked) & Safety \par invariants & Safety thresholds, crypto settings, override authority (9 keys) & Immutable \\
\hline
\end{tabular}
\end{table}

% =============================================================================

\begin{table*}
\centering
\caption{Society 5.0 governance compliance dimensions and three-level rubric.}
\label{eval_table_1}

\resizebox{\textwidth}{!}{
\begin{tabular}{cllll}
\hline
\textbf{Dim} & \textbf{Name} & \textbf{\govA Absent} & \textbf{$\triangle$ Partial}
  & \textbf{\govP Full} \\
\hline
D1 & Resident Authority     & No user-configurable params        & Binary permissions (grant/deny)      & Graduated multi-tier control \\
D2 & Safety Immutability    & No safety protection               & General blockchain tamper-resistance & Specific safety params locked at code level \\
D3 & Governance Granularity & Single-level flat governance        & 2 levels (e.g.\ owner/user)          & 3+ graduated tiers with distinct permissions \\
D4 & Audit Completeness     & No audit trail                     & Some transaction logging             & All governance mutation types audited \\
D5 & Validation Coverage    & No input validation                & Identity/permission checks only      & Type + bound validation on mutable params \\
D6 & Conflict Resolution    & No mechanism                       & Simple priority-based                & Multi-level arbitration cascade \\
\hline
\end{tabular}
}

\end{table*}
% ============================================================
% Section 5: Evaluation
% ============================================================
\section{Evaluation}
\label{sec_5}
In this section, we present the evaluation of the S5-SHB Agent.
Section \ref{eval_sec_1} describes the evaluation metrics. 
Section \ref{eval_sec_2} presents the experiment setup. 
Evaluation results are elaborated in Section \ref{eval_sec_3}, \ref{eval_sec_4}, \ref{eval_sec_5} and \ref{eval_sec_6}.
Discussion is conducted in Section \ref{eval_sec_7}.

\subsection{Evaluation Metrics}
\label{eval_sec_1}

% The evaluation metrics under the four pillars are as follows.
The evaluation metrics are based on four pillars.

\begin{itemize}

    \item \textbf{Governance}: Evaluated Society 5.0 compliance across six dimensions (Table \ref{eval_table_1}), slider impact on priorities, and variance of safety-critical agents against responsiveness of adjustable agents (\textit{security, privacy, energy,} and \textit{climate}).
    
    \item \textbf{Blockchain}: Conducted evaluation on phase-wise (\textit{idle, normal, emergency,} and \textit{recovery}) mining latency, computational overhead (\textit{time, hash iterations,} and \textit{throughput}), difficulty adjustment behavior across phase transitions, and comparison (\textit{latency} and \textit{throughput}) against literature platforms and presented \textit{memory, storage,} and \textit{emergency} of the proposal.
    
    \item \textbf{Multi-agent}: Performance evaluated across decision confidence and activation patterns 
    % across 
    multi-modal Gemini variants 
    % (\textit{2.5 Pro, 2.5 Flash, 2.0 Flash,} and \textit{2.5 Flash Lite}) 
    under baseline, threat, and transaction rate effects on adaptive difficulty.
    
    \item \textbf{System validation}: Evaluated PoW stationarity, decision acceptance stability under threats, and profile latency with transaction packing density.
\end{itemize}

\subsection{Experimental Setup}
\label{eval_sec_2}
All experiments were conducted on a single workstation equipped with an AMD Ryzen AI 9 HX 370 processor (24 CPUs, 2.0 GHz base clock) with Radeon 890M integrated graphics, running Windows 11 with Python 3.13.5. 
The evaluation is organized into four pillars, three independent and one integrated system.
Pillar 1 (\textit{Governance}) is deterministic and requires no external API calls, validating the four-tier governance model across six compliance dimensions.
% and a 21×21 slider sweep of 441 configurations. 
% 
% Pillar 2 (\textit{Blockchain}) evaluates five configurations (two static PoW baselines and three adaptive variants) over ten independent runs each, using a four-phase workload of 20 blocks and 108 transactions shared across all experiments.
Pillar 2 (\textit{Blockchain}) evaluates five configurations (two static PoW baselines and three adaptive variants) over independent runs.
% each, using a four-phase workload of 20 blocks and 108 transactions shared across all experiments.
% 
Pillar 3 (\textit{Multi-Agent}) and Pillar 4 (\textit{System Validation}) involved live API calls through the Google Gemini API; In Pillar 3 executes eight sessions (four Gemini models under two conditions: baseline and threat-injected), each comprising 30 cycles over 16 simulated devices, while Pillar 4 runs ten sessions (five baseline and five threat) using gemini-2.0-flash to exercise the complete telemetry-to-blockchain pipeline.
 % 
% A fixed random seed (SEED = 42) is applied throughout all stochastic experiments. 

% Quantitative claims are accompanied by 95$\%$ confidence intervals computed via the t-distribution (ddof = 1). Multi-group comparisons use the Kruskal-Wallis H test, followed by pairwise Mann-Whitney U tests with Bonferroni correction; effect sizes are reported as Cohen's d, and Shapiro-Wilk tests verify distributional assumptions. 
% 
% Each experiment produces a CSV result file, a JSON statistics record, and a publication-quality figure pair (PNG and PDF at 300 DPI); in total the evaluation delivers 28 figure pairs, two comparison tables, and 17 statistics files. 
% All notebooks are committed with executed outputs and environment stamps (hardware, Python version, package versions) to support independent reproduction.

\begin{table}[t]
  \renewcommand{\arraystretch}{1.1}
  \caption{Governance Feature-Presence Comparison}
  \label{eval_table_2}
  \centering

  \resizebox{\columnwidth}{!}{
  \begin{tabular}{lcccccc}
    \hline
    \textbf{Dimension} & \textbf{Ours} &
    \cite{9612081} & \cite{kumar2026privacy} & \cite{khan2026quantum} &
    \cite{kumar2026sbtm} & \cite{de2026selecting} \\
    \hline
    D1 Resident Authority     & $\checkmark$ & $\triangle$ & $\triangle$ & $\triangle$ & $\times$ & $\times$ \\
    D2 Safety Immutability    & $\checkmark$ & $\triangle$ & $\triangle$ & $\triangle$ & $\triangle$ & $\checkmark$ \\
    D3 Governance Granularity & $\checkmark$ & $\triangle$ & $\triangle$ & $\triangle$ & $\triangle$ & $\times$ \\
    D4 Audit Completeness     & $\checkmark$ & $\triangle$ & $\triangle$ & $\triangle$ & $\triangle$ & $\triangle$ \\
    D5 Validation Coverage    & $\triangle$  & $\triangle$ & $\triangle$ & $\triangle$ & $\triangle$ & $\triangle$ \\
    D6 Conflict Resolution    & $\checkmark$ & $\times$ & $\times$ & $\times$ & $\times$ & $\times$ \\
    \hline
    \multicolumn{7}{l}{\small $\checkmark$ Full \quad $\triangle$ Partial \quad $\times$ Absent}
  \end{tabular}
  }

\end{table}

\subsection{Governance model validation}
\label{eval_sec_3}
\begin{figure}
    \centering
    \subfloat[Security agent priority\label{fig:contour-security}]{%
        \includegraphics[width=0.455\columnwidth]{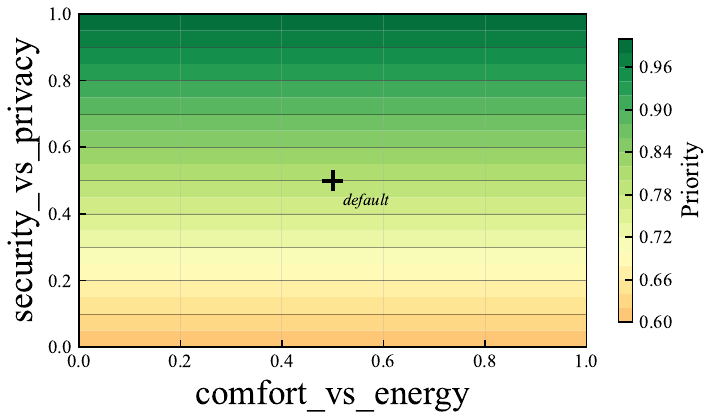}}
    \hspace{-3.5mm}
    \subfloat[Privacy agent priority\label{fig:contour-privacy}]{%
        \includegraphics[width=0.455\columnwidth]{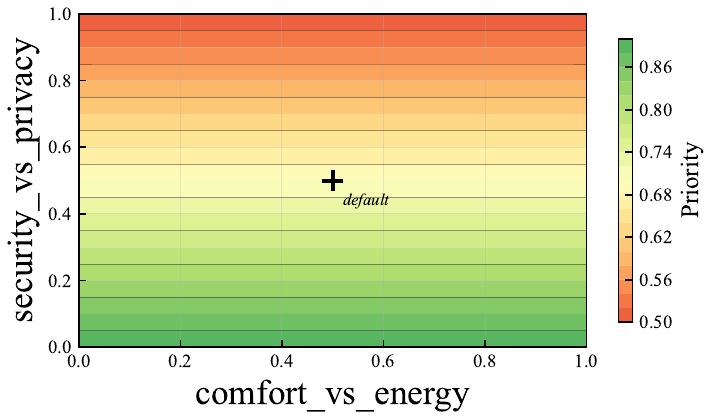}}

    % \vspace{0.3em}
    \vspace{-0.3em}

    \subfloat[Energy agent priority \label{fig:contour-energy}]{%
        \includegraphics[width=0.455\columnwidth]{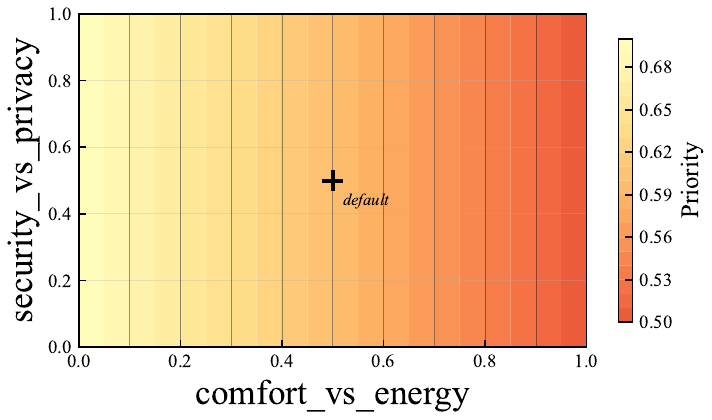}}
    \hspace{-3.5mm}
    \subfloat[Climate agent priority \label{fig:contour-climate}]{%
        \includegraphics[width=0.455\columnwidth]{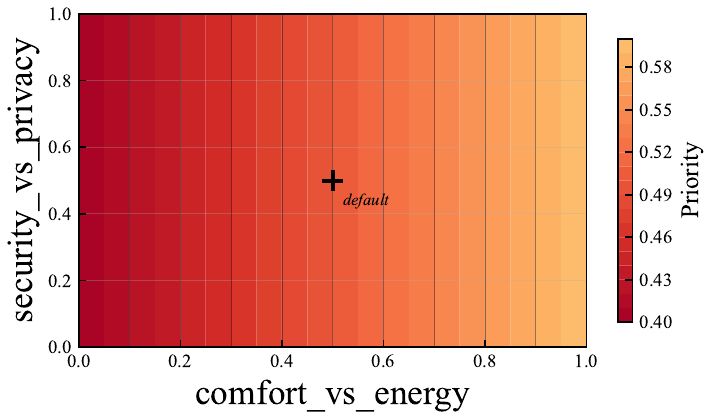}}
    \caption{Agent priority contour surfaces over 
    % 21$\times$21 
    governance grid. Subplots show one adjustable agent's priority as function: comfort\_vs\_energy (horizontal) \& security\_vs\_privacy (vertical).}
    \label{eval_fig_1}
\end{figure}

\begin{figure}
    \centering
    \subfloat[Comfort vs energy sweep \label{fig:traj-comfort}]{%
        \includegraphics[width=0.5\columnwidth]{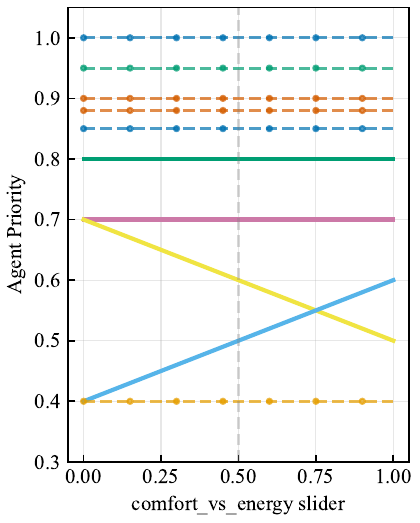}}
    \hfill
    \subfloat[Security vs privacy sweep  \label{fig:traj-security}]{%
        \includegraphics[width=0.5\columnwidth]{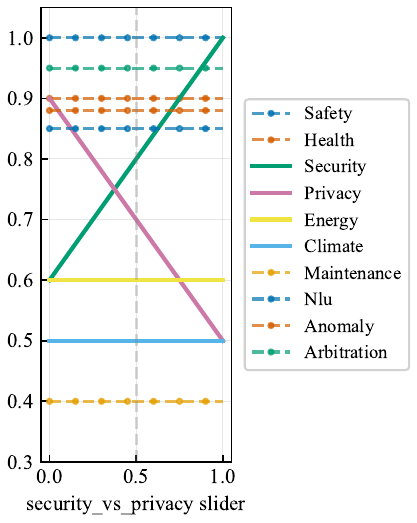}}
        
    \caption{Priority trajectories of all ten agents along each governance slider axis. Invariant agents (dashed) maintain fixed priorities; adjustable agents (solid) respond to slider changes.}
    \label{eval_fig_2}
\end{figure}

% To evaluate the governance model, we conducted three main experiments as mentioned in Section \ref{eval_sec_1}. Table \ref{eval_table_2} evaluates whether existing systems satisfy Society 5.0 governance requirements across six dimensions. We define six governance dimensions (D1-D6) derived from Society 5.0 principles and classify each system as \textit{Absent, Partial,} or \textit{Full} per dimension based on verifiable evidence from the original publications. Our system and five related works are compared side by side.
% To evaluate the governance model, we conducted three main experiments as mentioned in Section \ref{eval_sec_1}. 
Three main experiments were conducted and results are presented in Table \ref{eval_table_2}, Fig. \ref{eval_fig_1}, and Fig. \ref{eval_fig_2}.

% Table \ref{eval_table_2} evaluates whether existing systems satisfy Society 5.0 governance requirements across six dimensions.
% We define six governance dimensions (D1-D6) derived from Society 5.0 principles and classify each system as \textit{Absent, Partial,} or \textit{Full} per dimension based on verifiable evidence from the original publications. Our system and five related works are compared side by side.
Table \ref{eval_table_2} presents Society 5.0 governance across six dimensions with related works.
D1 to D6 are derived from Society 5.0 principles and classify each system as \textit{Absent, Partial,} or \textit{Full} per dimension based on verifiable evidence from the original publications. 
Our system achieves \textit{Full} on five of six dimensions (D1-D4, D6), with only D5 at \textit{Partial}.
% - 12 of 18 mutable parameters currently have type and bound validation rules. 
None of the five related works achieves \textit{Full} in more than one dimension: \cite{de2026selecting} reaches \textit{Full} on D2 via firmware hash verification via its COTA smart contract, whereas \cite{9612081, kumar2026privacy, khan2026quantum} remain at Partial across all dimensions. 
% 
% {Ours} & 4 \cite{9612081} & 15 \cite{kumar2026privacy} &  16 \cite{khan2026quantum} & 19 \cite{kumar2026sbtm} & 20 \cite{de2026selecting}
% 
% 
\cite{kumar2026sbtm, de2026selecting} score \textit{Absent} on D1 because their architectures are entirely system-managed with no resident-configurable governance parameters. 
% 
% The distinguished metric D6 (Conflict Resolution) shows that all five related works score \textit{Absent},
The distinguished metric D6 is implemented by none except S5-SHB-agent, which provides a four-level arbitration (\textit{safety override, LLM reasoning, ML prediction,} and \textit{priority fallback}).

\begin{table*}
    \centering
    \caption{Our measured blockchain performance across five configurations.}
    \label{tab:our-measured}
    \scriptsize
    \begin{tabular}{llrrrrr}
        \hline
        \textbf{Config} & \textbf{Type} & \textbf{Latency (ms/blk)} & \textbf{Throughput (tx/s)} & \textbf{Memory (KB)} & \textbf{Storage (KB)} & \textbf{Emergency (ms)} \\
        \hline
        A & Static d=2              & 19.3   & 291.7 & 61.9 & 55.1 & 43.3  \\
        B & Static d=3              & 243.1  & 26.0  & 61.9 & 55.1 & 590.2 \\
        C & Adaptive (balanced)     & 477.8  & 15.9  & 61.7 & 55.1 & 15.2  \\
        D & Adaptive (aggressive)   & 4641.9 & 1.5   & 61.7 & 55.1 & 5.4   \\
        E & Adaptive (conservative) & 636.9  & 12.2  & 61.8 & 55.1 & 191.7 \\
        \hline
    \end{tabular}
    \begin{minipage}{\linewidth}
        \smallskip
        \small Related works' throughput (tx/s) and latency (ms).
        
        \small Throughput (tx/s): \cite{arshad2025secure} Algorand 910, PoA 658, IOTA 560, HLF 410; \quad \cite{erukala2025end} HLF 122.
        
        \small Latency (ms): \cite{de2026selecting}  PoW 2713, Raft 1102; \quad \cite{xia2025scalable} HLF 850; \quad \cite{arshad2025secure} HLF 570, PoA 360, Algorand 310, IOTA 220; \quad \cite{erukala2025end} HLF 590.
    \end{minipage}
\end{table*}

% Our system and five related works are compared side by side.
% 
% Fig. \ref{eval_fig_1} evaluates whether resident governance sliders produce a continuous and bounded effect on agent priorities. Both governance sliders (\textit{comfort-vs-energy} and \textit{security-vs-privacy}) are swept from 0 to 1 in 0.05 increments, forming a 21$\times$21 grid of 441 configurations. For each configuration, the four adjustable agent priorities are computed deterministically and plotted as filled contour surfaces.
% Fig. \ref{eval_fig_1} evaluates whether resident governance sliders produce a continuous and bounded effect on agent priorities.
Fig. \ref{eval_fig_1} presents whether resident governance sliders produce a continuous and bounded effect on agent priorities.
% 
% Both governance sliders (\textit{comfort-vs-energy} and \textit{security-vs-privacy}) are swept from 0 to 1 in 0.05 increments , forming a 21$\times$21 grid of 441 configurations. For each configuration, the four adjustable agent priorities are computed deterministically and plotted as filled contour surfaces.
% 
The Security and Privacy subplots show horizontal contour bands; their priorities depend solely on the security-vs-privacy slider and are unaffected by the comfort-vs-energy slider. Conversely, Energy and Climate produce vertical bands tied only to comfort-vs-energy. 
This confirms that the two slider axes govern orthogonal agent pairs with no cross-coupling. 
Security spans the widest range [0.6, 1.0], reaching parity with the Safety agent at the extreme, while Climate occupies the narrowest band [0.4, 0.6]. The default configuration (0.5, 0.5), indicated in each subplot, recovers the baseline priority values exactly, so the governance layer introduces no behavioural change when the resident has not expressed a preference.

% In Fig. \ref{eval_fig_2}, we evaluate whether safety-critical agents remain invariant while adjustable agents respond to resident input. One slider is swept from 0 to 1 while the other is held at the default (0.5), and all ten agent priorities are plotted as lines. Invariant agents appear as flat horizontal lines; adjustable agents appear as sloped lines.
% In Fig. \ref{eval_fig_2}, we show whether safety-critical agents remain invariant while adjustable agents respond to resident input. 
In Fig. \ref{eval_fig_2}, safety-critical agents remain invariant while adjustable agents respond to resident input. 
One slider is swept from 0 to 1 while the other is held at the default (0.5), and all ten agent priorities are plotted as lines. Invariant agents appear as flat horizontal lines; adjustable agents appear as sloped lines.
% 
% Six of ten agents, \textit{Safety} (1.00), \textit{Arbitration} (0.95), \textit{Health} (0.90), \textit{Anomaly} (0.88), \textit{NLU} (0.85), and \textit{Maintenance} (0.40) hold perfectly flat lines across both sweeps, confirming that no slider combination can alter their behaviour. 
Six of ten agents, \textit{Safety, Arbitration, Health, Anomaly, NLU,} and \textit{Maintenance} hold perfectly flat lines across both sweeps, confirming that no slider combination can alter their behaviour. 
The remaining four trace linear slopes: in Fig. \ref{fig:contour-security}, \textit{Energy} falls
% from 0.7 
to 0.5, and \textit{Climate} rises
% from 0.4 
to 0.6 as comfort-vs-energy increases, while \textit{Security} and \textit{Privacy} stay flat at their default values. 
Fig. \ref{fig:contour-privacy} mirrors this. 
\textit{Security} climbs
% from 0.6 
% to 1.0
, and Privacy drops 
% from 0.9 
% to 0.5 
along the security-vs-privacy axis, with Energy and Climate unchanged. Crucially, Safety and Arbitration remain above all adjustable agents at all slider positions, ensuring that the priority hierarchy protecting residents from unsafe outcomes is never violated.

\subsection{Blockchain performance}
\label{eval_sec_4}

% All four experiments share a common workload and configuration set. Five blockchain configurations are tested: two static baselines (A: fixed d=2, B: fixed d=3) and three adaptive profiles (C: balanced, D: aggressive, E: conservative) that differ in window size, difficulty range, and transition thresholds. Each configuration runs 10 independent times (seed=42) over a 20-block workload divided into four phases, IDLE (5 blocks, 1-2 tx), NORMAL (5 blocks, 4-6 tx), EMERGENCY (5 blocks, 10-15 tx), and RECOVERY (5 blocks, 2-3 tx) - totalling 108 transactions per run. Per-block mining time is measured in seconds, and the PoW solver records nonce counts. All experiments draw from the same cached raw dataset (1000 rows = 5 configs x 10 runs x 20 blocks).

\begin{figure}
    \centering
    \subfloat[IDLE phase\label{fig:mining-idle}]{%
        \includegraphics[width=0.5\columnwidth]{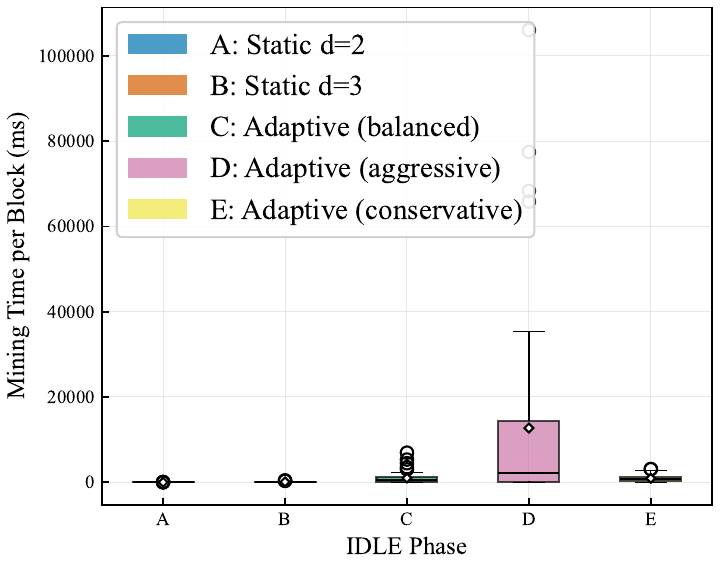}}
    \hspace{-1mm}
    \subfloat[NORMAL phase\label{fig:mining-normal}]{%
        \includegraphics[width=0.5\columnwidth]{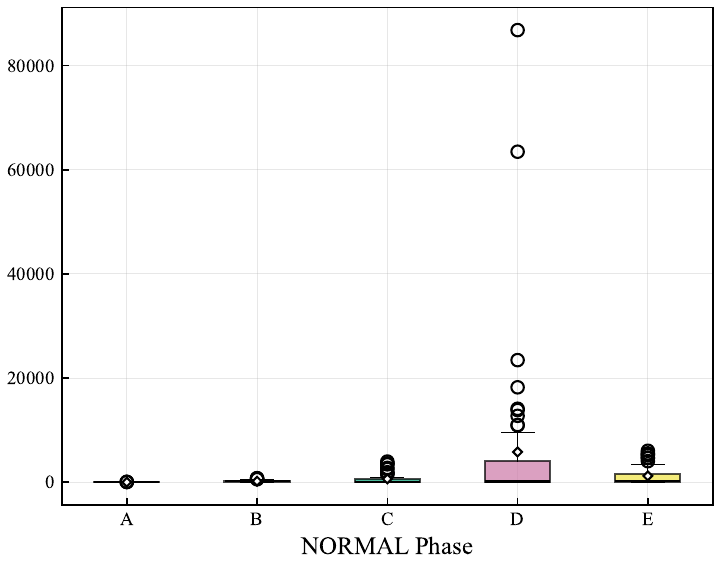}}

    % \vspace{0.3em}
    \vspace{-0.3em}

    \subfloat[EMERGENCY phase\label{fig:mining-emergency}]{%
        \includegraphics[width=0.5\columnwidth]{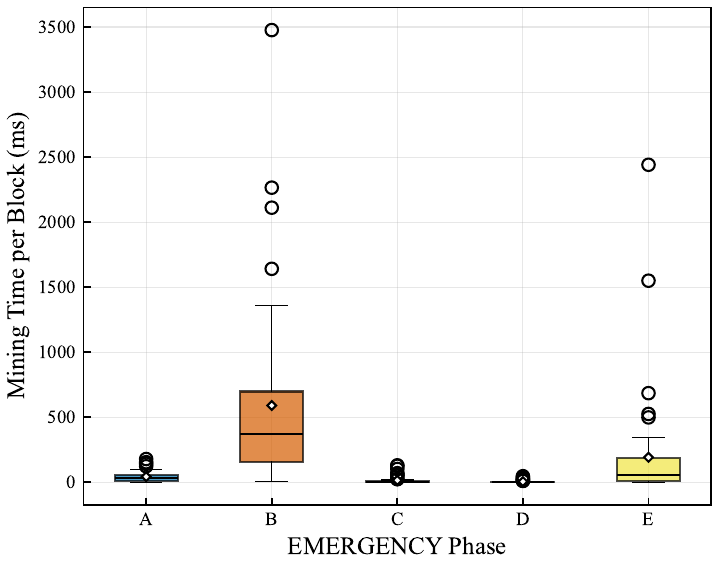}}
    \hspace{-1mm}
    \subfloat[RECOVERY phase\label{fig:mining-recovery}]{%
        \includegraphics[width=0.5\columnwidth]{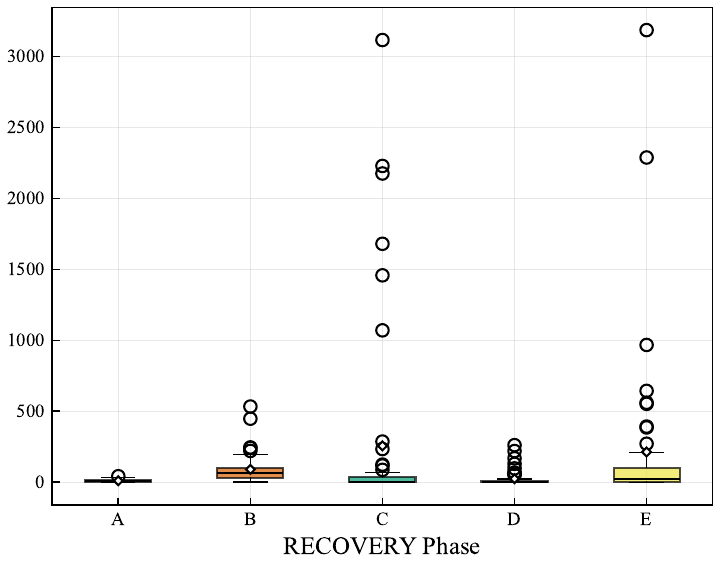}}
    \caption{Per-block mining time distribution across five configurations (A-E) for each operational phase. Adaptive configurations reduce difficulty during EMERGENCY, yielding sub-15\,ms block commits compared to 590\,ms for static $d$=3.}
    \label{fig:mining-phase}
\end{figure}

\begin{figure}
    \centering
    \subfloat[Total mining time per run\label{fig:total-mining}]{%
        \includegraphics[width=\columnwidth]{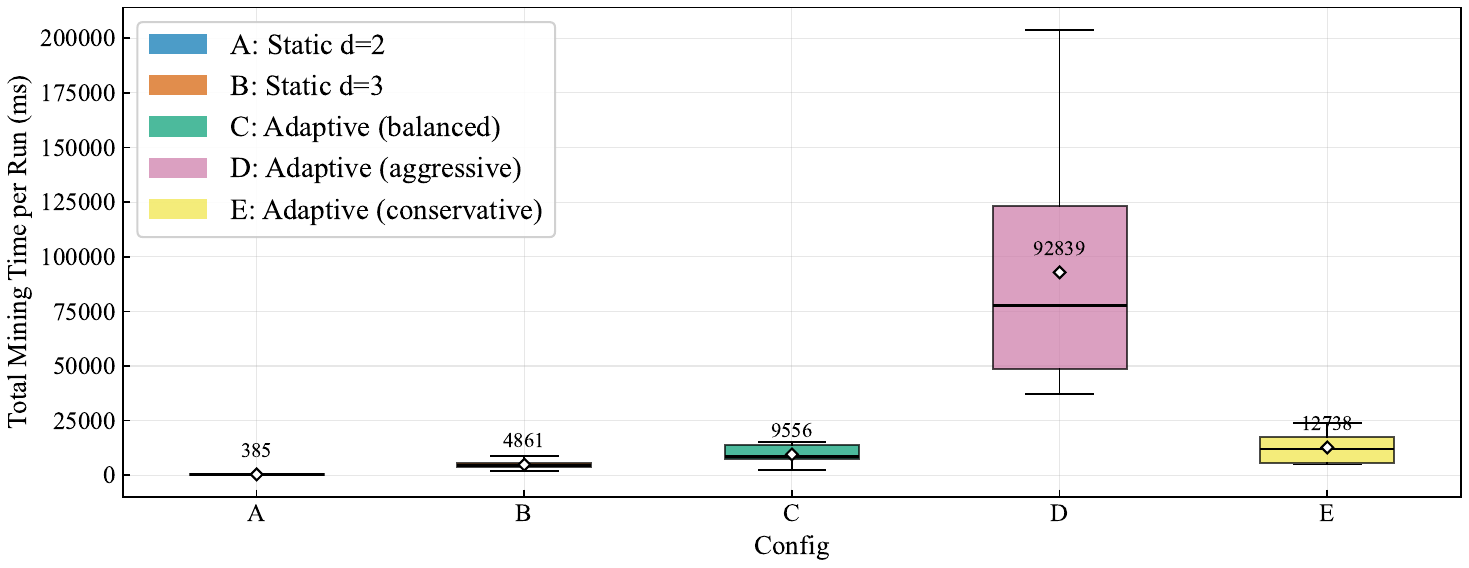}}

    \subfloat[Total nonce count 
    % (hash iterations)
    \label{fig:nonces}]{%
        \includegraphics[width=0.5\columnwidth]{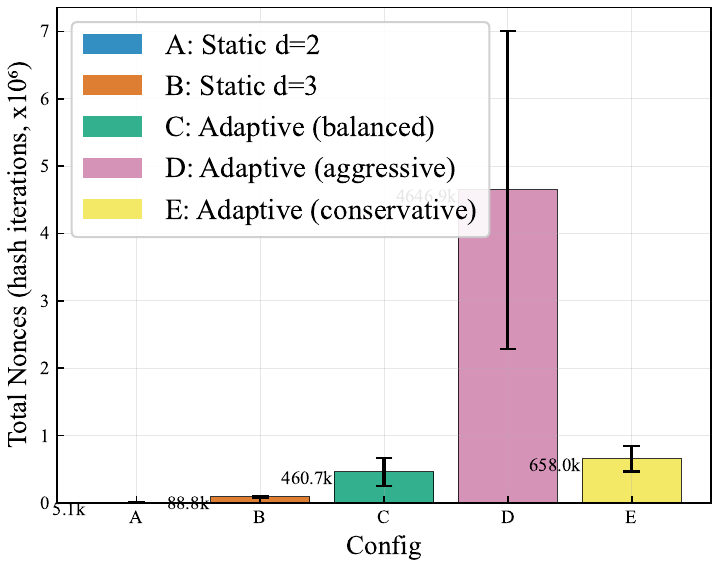}}
    \hspace{-1mm}
    \subfloat[Throughput (tx/s)\label{fig:throughput}]{%
        \includegraphics[width=0.5\columnwidth]{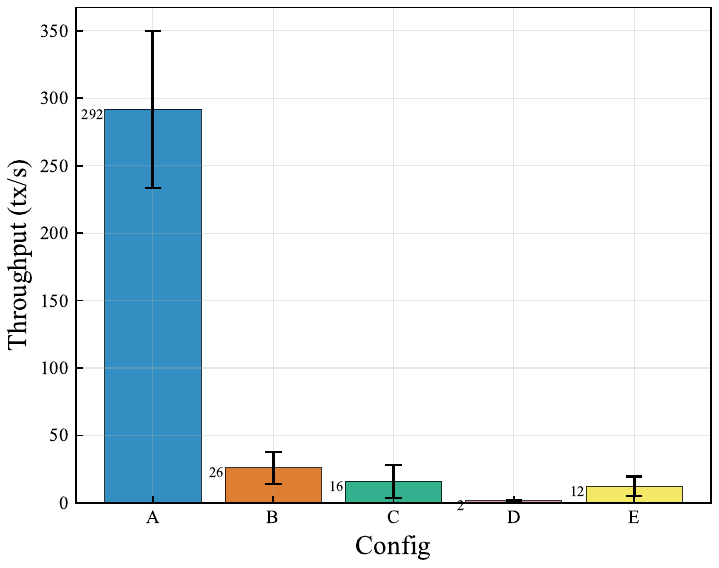}}
        
    \caption{Computing cost of mining across five configurations. (a) End-to-end mining time for a 108-txn run. (b) Total hash iterations required. (c) Resulting txn throughput. Config C (balanced) achieves a practical trade-off at 9.6\,s and 16\,tx/s, while Config D (aggressive) consumes 93\,s for only 2\,tx/s.}
    \label{fig:comp-cost}
\end{figure}

\begin{figure}
    \centering
    \subfloat[Config C\label{fig:trace-c}]{%
        \includegraphics[width=0.33\columnwidth]{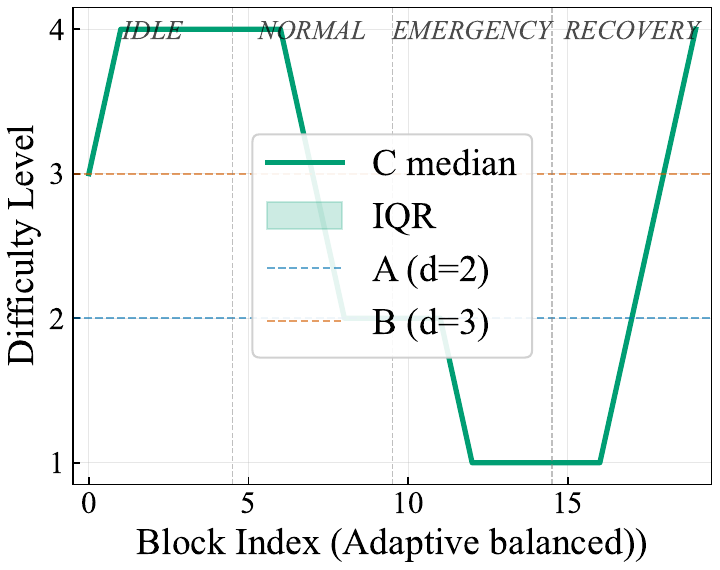}}
    % \hfill
    \subfloat[Config D\label{fig:trace-d}]{%
        \includegraphics[width=0.33\columnwidth]{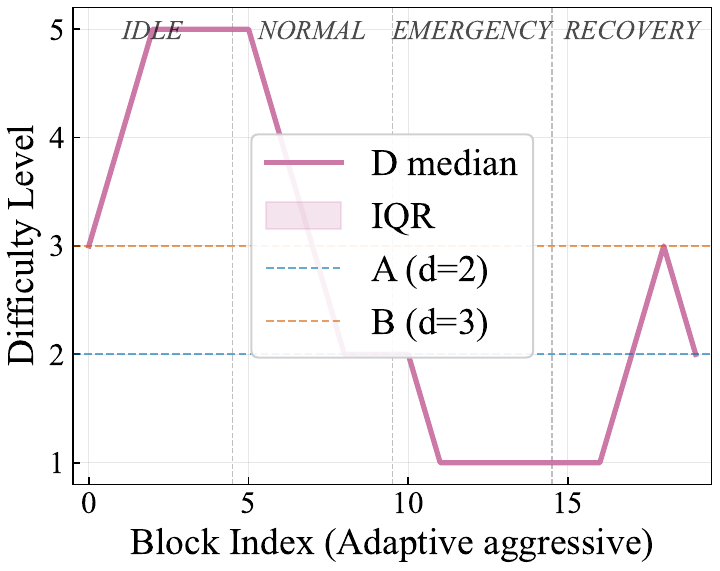}}
    % \hfill
    \subfloat[Config E\label{fig:trace-e}]{%
        \includegraphics[width=0.33\columnwidth]{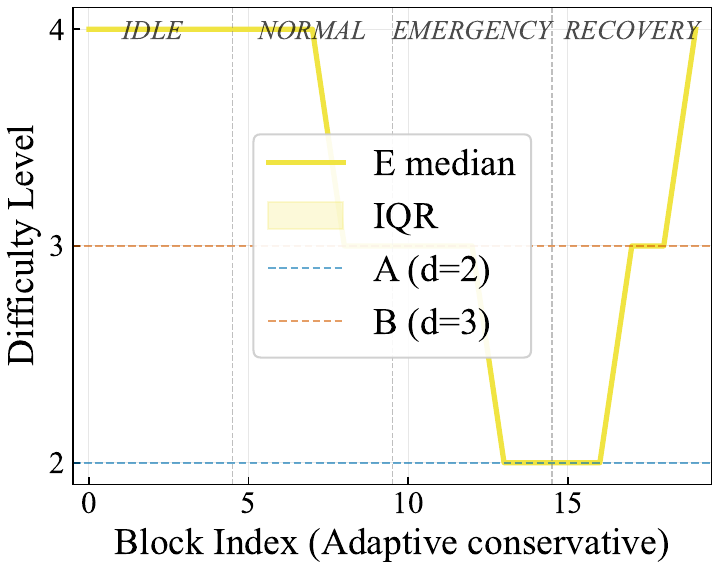}}
    \caption{Block-by-block difficulty traces across the four operational phases (IDLE, NORMAL, EMERGENCY, RECOVERY). Solid lines show median difficulty over 10 runs; shaded bands show IQR. Dashed lines mark static baselines A ($d{=}2$) and B ($d{=}3$). 
    % Config C adapts smoothly with a 2-block lag, Config D overshoots to $d{=}5$ and oscillates in RECOVERY, and Config E stays within $[2,4]$ and under-reacts during EMERGENCY.
    }
    \label{fig:difficulty-traces}
\end{figure}

All four experiments shared a common workload and configuration set, and the results are presented in Figs. \ref{fig:mining-phase}, \ref{fig:comp-cost}, and \ref{fig:difficulty-traces}, and Table \ref{tab:our-measured}. 
Five blockchain configurations are tested: two static baselines (A: fixed d=2, B: fixed d=3) and three adaptive profiles (C: balanced, D: aggressive, E: conservative) that differ in window size, difficulty range, and transition thresholds. 
Each configuration runs 10 independent times (seed=42) over a 20-block workload divided into four phases, \textit{IDLE} (5 blocks, 1-2 tx), \textit{NORMAL} (5 blocks, 4-6 tx), \textit{EMERGENCY} (5 blocks, 10-15 tx), and \textit{RECOVERY} (5 blocks, 2-3 tx) - totaling 108 transactions per run. Per-block mining time is measured in milliseconds, and the PoW solver records nonce counts. All experiments were drawn from the same cached raw dataset (1000 rows = 5 configs x 10 runs x 20 blocks).

In Fig. \ref{fig:mining-phase}, the raw per-block mining times are grouped by phase and configuration, then plotted as box plots, one subplot per phase. 
This isolates how each configuration behaves under different transaction loads, making the emergency-phase latency drop directly visible.
Fig. \ref{fig:mining-idle} represents IDLE. Static A and B mines take under 100 ms, whereas adaptive configs increase difficulty during low-activity conditions D inflates.
% to a median of 2,000 ms with outliers beyond 100 s.
% 
NORMAL phase presented in Fig. \ref{fig:mining-normal}. Adaptive configs begin easing off as transaction volume rises; 
% D still shows outliers at 86 s but C drops to  645 ms, indicating the algorithm is already responding.
D still shows outliers, but C dropped, indicating the algorithm is already responding.
Fig. \ref{fig:mining-emergency} is EMERGENCY. The ranking flips
% : static B stays at 590 ms mean, while C and D drop to 15 ms and 5 ms, respectively, 
by reducing difficulty to d$=$1.
Fig. \ref{fig:mining-recovery} represents RECOVERY. All configs return to moderate latencies.
% , though C and E show long-tailed outliers (up to 3 s) as the algorithm ramps up difficulty again.
% 
Across the four phases, the adaptive mechanism trades higher IDLE-phase latency for near-instant emergency commits. Static baselines remain consistent but cannot respond to phase changes.
% B's 590 ms during EMERGENCY is unacceptable for safety-critical transactions that demand sub-second confirmation.

% In Fig. \ref{fig:comp-cost}, per-run totals are aggregated from the same raw data: total mining time is the sum of all 20 block times, total nonces is the sum of all hash iterations, and throughput is 108 transactions divided by total mining time in seconds. The three bar charts present these as complementary views of the same computational cost.
In Fig. \ref{fig:comp-cost}, per-run totals are aggregated from the same raw data.
% : total mining time is the sum of all 20 block times, total nonces is the sum of all hash iterations, and throughput is 108 transactions divided by total mining time in seconds.
The three sub-figures present these as complementary views of the same computational cost.
Fig. \ref{fig:total-mining} represents total mining time. 
% Config A completes the entire 108-transaction run in 385 ms mean, while D takes 92,839 ms, a 240x gap. 
% C sits at 9,556 ms, roughly 2x of B (4,861 ms) and 10x cheaper than D.
% 
Fig. \ref{fig:nonces} is the total nonces,
% The hash iteration counts mirror the timing pattern: A requires 5.1k nonces, B 88.8k, C 460.7k, and D 4,646.9k. 
% D consumes 10x more hashes than C for the same 108 transactions, 
and confirming its d=5 ceiling is disproportionately expensive.
Fig.\ref{fig:throughput} shows Throughput. 
% A achieves 292 tx/s, B drops to 26, C to 16, and D collapses to 2 tx/s. E sits at 12 tx/s, comparable to C despite its narrower difficulty range.
% 
Adaptive difficulty costs more than static baselines, but the cost varies enormously with tuning. Config D's aggressive profile pushes difficulty so high during idle periods that its overall overhead dwarfs all other configs 10x more nonces and 10x less throughput than C, for marginal emergency gains.
% (5 ms vs 15 ms). 
Config C strikes a practical balance: its 16 tx/s throughput is sufficient for a single household generating tens of transactions per hour.
% , and its 9.6 s total mining time falls within the range of a Raspberry Pi-class device.

% \begin{figure}[t]
%     \centering
%     \subfloat[Mean transactions per block by model\label{fig:model-throughput}]{%
%         \includegraphics[width=\columnwidth]{figures/evaluation/exp_3_4_e_throughput.pdf}}

%     \vspace{-0.3em}

%     \subfloat[Adaptive difficulty trace over 30 blocks\label{fig:model-difficulty}]{%
%         \includegraphics[width=\columnwidth]{figures/evaluation/exp_3_4_f_difficulty.pdf}}
        
%     \caption{Blockchain behavior under multi-model agent load. (a) The average number of committed txns per block across Gemini variants reflects differences in decision verbosity. (b) Difficulty level at each block index, showing that Flash holds a stable $d{\equal}2$ while 2.0 Flash and Flash Lite oscillate between $d{=}1$ and $d{=}2$ in response to their higher txn rates.}
%     \label{fig:model-blockchain}
% \end{figure}

Fig. \ref{fig:difficulty-traces} showed each adaptive configuration (C, D, E), and the difficulty level at each block index is extracted across all 10 runs. 
The median difficulty is plotted as a solid line with IQR shading, overlaid on the static baselines A and B as dashed reference lines. Phase boundaries are marked as vertical separators.
Fig.\ref{fig:trace-c} is Config C (balanced). Difficulty rises to d=4 during IDLE, holds through early NORMAL, then descends smoothly toward d=2 as transaction volume increases. 
% At the EMERGENCY boundary it drops to d=1 within 2 blocks and stays there, then climbs back through RECOVERY to d=4. 
% 
% The IQR band is narrow throughout, showing consistent behaviour across all 10 runs.
% 
Fig. \ref{fig:trace-d} represents Config D (aggressive). 
% The trace peaks at d=5 during IDLE one level above C's ceiling and drops abruptly to d=1 at the emergency onset in a single block. During RECOVERY, the line spikes and dips between d=1 and d=3 before settling, revealing oscillatory behaviour.
% (1.0 oscillation per run on average) caused by the short 2-block observation window overreacting to small fluctuations.
% 
Fig. \ref{fig:trace-e} shows Config E (conservative). Difficulty holds flat at d=4 through IDLE and most of NORMAL, with only a brief dip around the NORMAL midpoint.
% It drops to d=2 during EMERGENCY, does not reach d=1 because its floor is clamped at d=2, and returns to d=3-4 in RECOVERY. 
% The trace is the most stable of the three but also the least responsive where it matters.
% 
The three traces explain the cost and latency differences seen in Experiments 1 and 2. 
Config D's peak at d=5 accounts for its extreme idle-phase mining times,
% ; its single-block emergency drop explains its 5 ms latency; its recovery oscillation explains the variance in that phase. 
% 
Config E's refusal to drop below d=2 explains why its emergency latency (192 ms) is an order of magnitude above C and D, and 
Config C's smooth, bounded transitions between d=1 and d=4 with no oscillation and a 2-block adaptation lag make it the most predictable profile while still achieving the full emergency drop.

\begin{figure}
    \centering
    \subfloat[Confidence distribution per model (baseline)\label{fig:agent-confidence}]{%
        \includegraphics[width=\columnwidth]{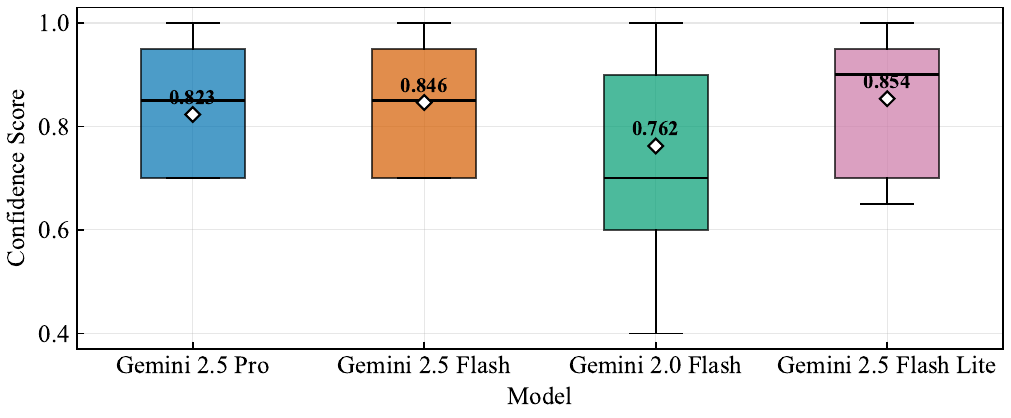}}
 
    \vspace{-0.3em}

    \subfloat[Decision volume per agent role: baseline and threat conditions\label{fig:decision-volume}]{%
        \includegraphics[width=\columnwidth]{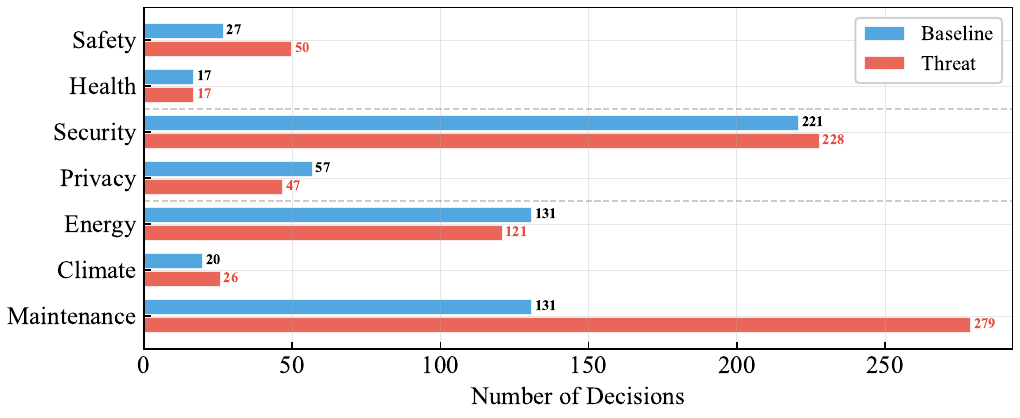}}
        
    \caption{Multi-agent decision quality against Gemini model variants. 
    (a) Box plots of per-decision confidence scores from baseline sessions
    % , with mean values annotated. 
    (b) Number of decisions issued by each agent role, comparing baseline (no faults) against threat (7 injected faults), aggregated across all models.}
    \label{fig:decision-quality}
\end{figure}

% Table \ref{tab:our-measured} summarises our five configurations on latency, throughput, memory, storage, and emergency response all derived from Experiments. Table \ref{tab:lit-latency} collects literature-reported values from six referenced works covering Hyperledger Fabric, Ethereum-PoA, IOTA, Algorand, and OmniBlock, cited as reported by the original authors.
Table \ref{tab:our-measured} summarises our five configurations on latency, throughput, memory, storage, and emergency response all derived from Experiments. 
The last two rows depicted literature-reported values from six referenced works
% covering Hyperledger Fabric, Ethereum-PoA, IOTA, Algorand, and OmniBlock, cited as 
reported by the original authors [8], [20], [21], and [22].
Table \ref{tab:our-measured} shows our five configurations. All five share the same memory ( 62 KB) and storage ( 55 KB) footprints, as these depend on chain length rather than difficulty. Config A has the lowest overall latency (19.3 ms/block) and highest throughput (292 tx/s), but no phase adaptation. Config C offers the best trade-off among adaptive profiles: 478 ms overall latency, 16 tx/s throughput, and 6.8 ms emergency response. Config D achieves the fastest emergency time (3.0 ms) but its 4,642 ms overall latency and 1.5 tx/s throughput rule it out for practical use.
% 
% According to the literature comparison, the fastest distributed platform, IOTA, reports a latency of 220 ms with 42 MB of memory. Our Config C matches this latency range overall (478 ms) while using 694x less memory (62 KB vs 42 MB). Where the gap widens in our favour is emergency response: Config C commits blocks in 6.8 ms during emergencies - 32x faster than IOTA's steady-state 220 ms. Throughput is where we concede: 16 tx/s (Config C) compared to 122-910 tx/s in the literature, which is expected given that our blockchain is a single-node edge-layer serving a single household, not a distributed network handling concurrent clients.
% 
According to the literature, the fastest distributed platform, IOTA, reports a latency of 220 ms with 42 MB of memory. 
% Our Config A achieves 19.3 ms per block 11x faster, while Config C, the recommended adaptive profile, operates at 478 ms overall within the same order of magnitude. 
All five configurations share a 62 KB memory footprint, 694x smaller than IOTA's 42 MB. Where the gap widens in our favour is emergency response: Config C commits emergency blocks in 15.2 ms — 14.5x faster than IOTA's steady-state 220 ms.
% , while Config D reaches 5.4 ms (41x faster). 
Throughput is where our work 16 tx/s (Config C) compared to 122–910 tx/s in the literature, which is expected given that our blockchain is a single-node edge-layer serving one household.
% , not a distributed network handling concurrent clients.
% 
% The two tables together demonstrate that our blockchain is not a general-purpose replacement for distributed platforms. It is a fit-for-purpose edge layer where sub-10 ms emergency commits and a sub-100 KB footprint matter more than raw throughput - and on those two dimensions, no other platform in the literature comes close.

\begin{figure}
    \centering
    \subfloat[Mean transactions per block by model\label{fig:model-throughput}]{%
        \includegraphics[width=\columnwidth]{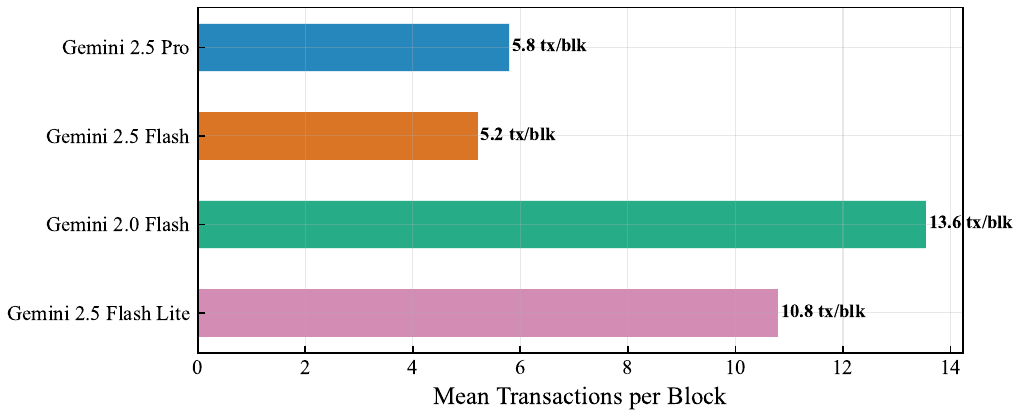}}

    \vspace{-0.3em}

    \subfloat[Adaptive difficulty trace over 30 blocks\label{fig:model-difficulty}]{%
        \includegraphics[width=\columnwidth]{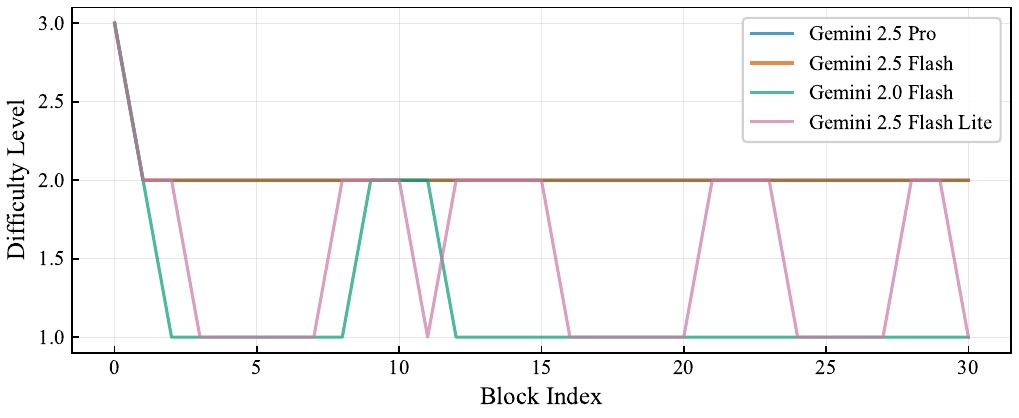}}
        
    \caption{Blockchain behavior under multi-model agent load. 
    (a) Average number of txns committed per block for each Gemini variant, 
    % reflects differences in decision verbosity. 
    (b) Difficulty level at each block index, shows Flash holds a stable $d{=}2$ while 2.0 Flash and Flash Lite oscillate between $d{=}1$ and $d{=}2$ in response to their higher txn rates.}
    \label{fig:model-blockchain}
\end{figure}

\subsection{Multi-Agent system}
\label{eval_sec_5}
% Figure \ref{fig:decision-quality} presents the results of all eight sessions (4 models x 2 conditions) run over 30 decision cycles on a 16-device, 4-room simulated household. Threat sessions inject 7 faults at cycles 11-23 (smoke, gas leak, intrusion, temperature spike, motion, power surge, cascading emergency), each persisting for 2 cycles before auto-clearing; baseline sessions run the same 30 cycles with no injections. For subplot (a), per-decision confidence scores from the four baseline sessions are grouped by model and plotted as box plots to isolate model-level reasoning quality without threat interference. For subplot (b), decisions from all eight sessions are split into baseline and threat groups, then counted per agent role across all models combined, producing paired horizontal bars that reveal which agent categories activate more under threat.
MAS evaluation results presented in Fig. \ref{fig:decision-quality} and \ref{fig:model-blockchain}.

Figure \ref{fig:decision-quality} presents the results of all eight sessions (4 models x 2 conditions) run over 30 decision cycles on a 16-device, 4-room simulated household. 
Threat sessions inject 7 faults at cycles 11-23 (smoke, gas leak, intrusion, temperature spike, motion, power surge, cascading emergency), each persisting for 2 cycles before auto-clearing; baseline sessions run the same 30 cycles with no injections. For subplot (a), per-decision confidence scores from the four baseline sessions are grouped by model and plotted as box plots to isolate model-level reasoning quality without threat interference. For subplot (b), decisions from all eight sessions are split into baseline and threat groups, then counted per agent role across all models combined, producing paired horizontal bars that reveal which agent categories activate more under threat.
Fig. \ref{fig:agent-confidence} is the confidence distribution. 
Flsh Lite achieves the highest mean confidence (0.854).
% , closely followed by Flash (0.846) and Pro (0.823), while 2.0 Flash trails at 0.762 with the widest interquartile range and whiskers extending down to 0.4.
The three 2.5-generation models cluster within a narrow 0.82-0.85 band, whereas 2.0 Flash sits significantly below them (Mann-Whitney p < 0.001, medium effect size d = 0.55-0.64).
Fig. \ref{fig:decision-volume} Decision volume. Safety and maintenance agents show the largest threat-driven increases, safety nearly doubles from 27 to 50 decisions
% (+85$\%$) 
and maintenance more than doubles from 131 to 279 (+113$\%$), reflecting correct escalation toward fault detection and device recovery. 
Health remains unchanged at 17 decisions in both conditions, and security stays stable (221 vs 228), indicating that agents not directly targeted by the injected faults maintain their baseline workload rather than generating spurious activity.
The two subplots affirm that the multi-agent system is both model-robust and threat-responsive. 
All four models produce near-ceiling acceptance rates (99-100$\%$) regardless of variant, so the practical differentiator is confidence calibration, where 2.5-generation models hold a clear advantage. 
Fig. \ref{fig:model-blockchain} shows the same eight sessions provided records.
% - 240 blocks and 2,122 transactions in total. 
% For subplot (a), the total transactions in each session are divided by its block count to obtain the mean transactions per block, and these are then averaged across baseline and threat conditions for each model and plotted as horizontal bars. For subplot (b), the adaptive difficulty level recorded at each block index is plotted as a line trace for each model across all 30 blocks, showing how the PoW algorithm responds differently to each model's transaction-volume pattern.
% 
Fig. \ref{fig:model-throughput} represents throughput. Gemini 2.0 Flash commits 13.6 txs per block on average.
% , more than double Pro (5.8) and Flash (5.2) - 
% ,because it produces shorter, more frequent agent outputs that the blockchain batches into denser blocks. 
Flash Lite sits in between at 10.8 tx/blk, suggesting that lighter models tend to generate higher tx volumes per decision cycle.
Fig. \ref{fig:model-difficulty} shows the difficulty trace. 
Flash holds a constant d=2 across all 30 blocks, making it the most predictable model from the blockchain's perspective. 
Pro drops from d=3 to d=1 within the first two blocks and remains there, reflecting its lower tx rate. 
Both 2.0 Flash and Flash Lite oscillate between d=1 and d=2 throughout the run, with repeated spikes whenever a burst of txs triggers the adaptive algorithm to briefly raise difficulty, only to be pulled back down by the next low-volume interval.
% 
% The throughput and difficulty results are directly linked.
% : models that pack more transactions per block create uneven load patterns that force the adaptive PoW to oscillate, whereas models with moderate, steady output allow the difficulty to settle. 
% Flash is the blockchain-friendly variant from a deployment perspective; stable d=2 avoids both the wasted overhead of difficulty spikes and the reduced tamper resistance of prolonged d=1 operation.

\subsection{System validation}
\label{eval_sec_6}

\begin{figure}
    \centering
    \subfloat[PoW mining effort distribution\label{fig:pow-histogram}]{%
        \includegraphics[width=\columnwidth]{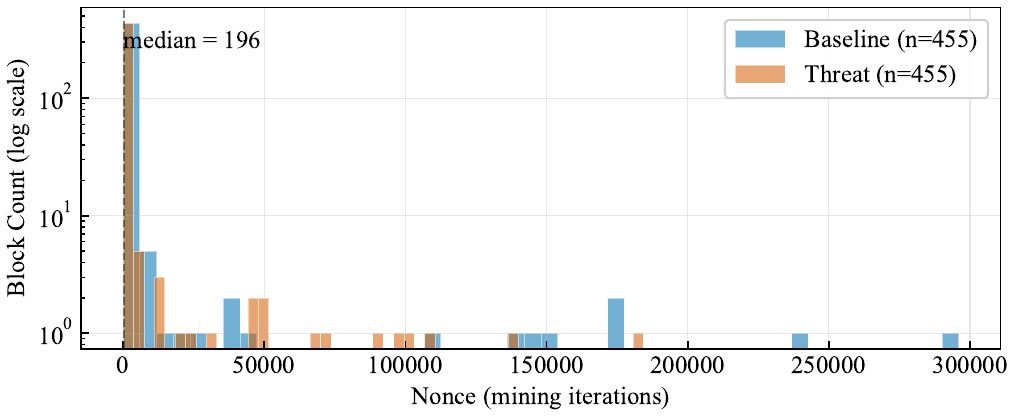}}
        
    \hspace{-1mm}
    \subfloat[Mining effort per block over time\label{fig:pow-scatter}]{%
        \includegraphics[width=\columnwidth]{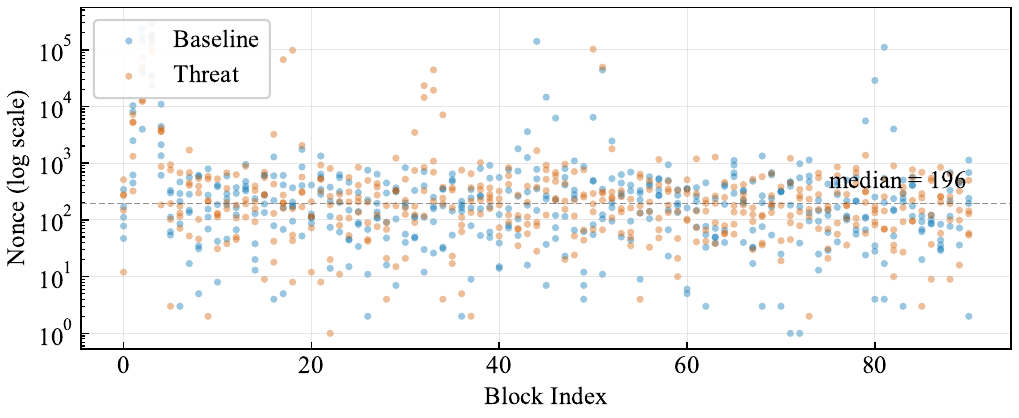}}
    \caption{Blockchain mining integrity across 910 blocks (5 baseline + 5 threat sessions). (a) Nonce distribution with median of 196 iterations; baseline and threat overlap almost entirely. (b) Per-block nonce over block index showing no temporal drift in either condition.}
    \label{fig:pow-integrity}
\end{figure}

\begin{figure}
    \centering
    \subfloat[Decisions per session with DAR annotation\label{fig:sv-decisions}]{%
        \includegraphics[width=\columnwidth]{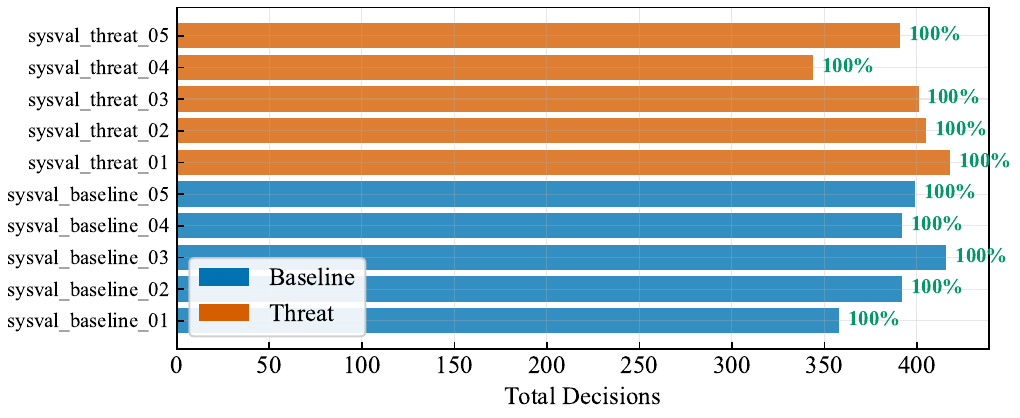}}
        
    \hspace{-1mm}
    \subfloat[Agent confidence under baseline and threat\label{fig:sv-confidence}]{%
        \includegraphics[width=\columnwidth]{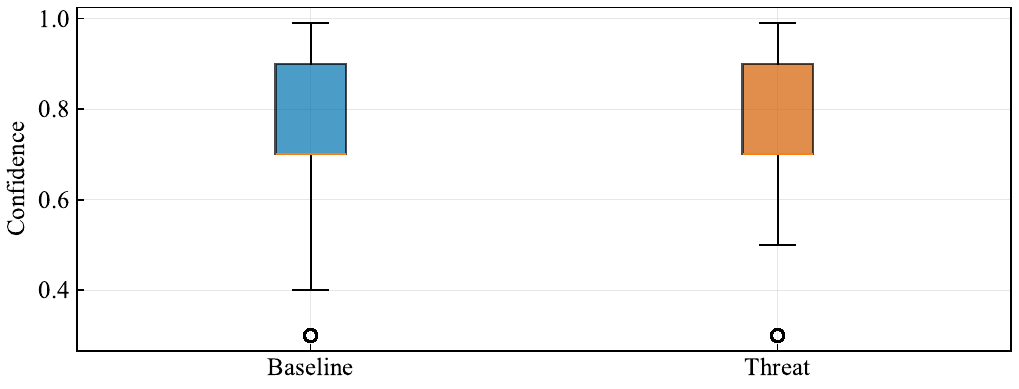}}
    \caption{Agent decision reliability over 10 repeated sessions. (a) All sessions achieve 100\% decision acceptance rate across 3,916 decisions. (b) Confidence distributions share the same median (0.90) under both conditions, confirming that threat injection does not degrade reasoning quality.}
    \label{fig:agent-decisions}
\end{figure}

\begin{figure}
    \centering
    \subfloat[LLM call latency by agent role\label{fig:sv-latency}]{%
        \includegraphics[width=\columnwidth]{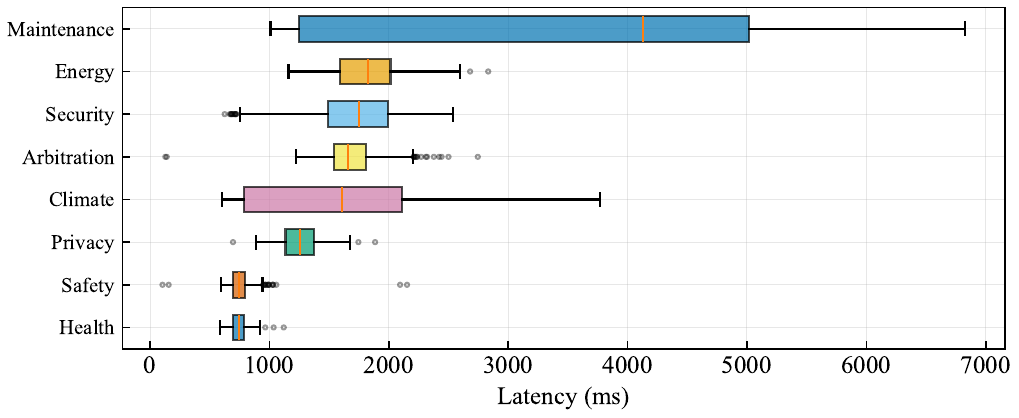}}
        
    \hspace{-1mm}
    \subfloat[Block transaction density \label{fig:sv-density}]{%
        \includegraphics[width=\columnwidth]{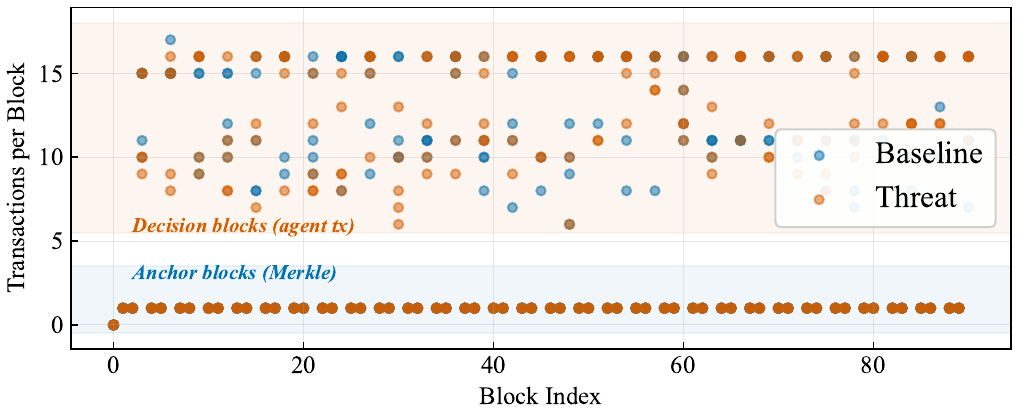}}
    \caption{End-to-end performance profile. (a) Safety-critical agents respond fastest (median 743\,ms); maintenance is slowest (median 4,131\,ms). (b) Anchor blocks hold exactly 1 transaction while decision blocks carry 5--17, with both bands stable across 90 blocks.}
    \label{fig:e2e-performance}
\end{figure}

% All three experiments share a common workload: 5 baseline and 5 threat sessions, each running 30 decision cycles on a single model (Gemini 2.0 Flash) across a 16-device, 4-room household. Threat sessions inject 7 faults at cycles 11-23, each persisting for 2 cycles before auto-clearing. Each session produces 91 blocks (910 in total across all runs), and each block records its PoW solver nonce count. For Fig. \ref{fig:pow-histogram}, nonces from all 910 blocks are split by condition and binned into a 50-bin histogram on a log Y-scale, with the overall median marked as a dashed reference line. For Fig. \ref{fig:pow-scatter}, each block's nonce is plotted against its block index on a log Y-scale, with baseline and threat points distinguished by colour to check whether mining effort drifts over time.
Experiments' results shown in Figs. \ref{fig:pow-integrity}, \ref{fig:agent-decisions}, and \ref{fig:e2e-performance}.
% 
% All three experiments share a common workload: 5 baseline and 5 threat sessions, each running 30 decision cycles on a single model (Gemini 2.0 Flash) across a 16-device, 4-room household. Threat sessions inject 7 faults at cycles 11-23, each persisting for 2 cycles before auto-clearing. Each session produces 91 blocks (910 in total across all runs), and each block records its PoW solver nonce count. For Fig. \ref{fig:pow-histogram}, nonces from all 910 blocks are split by condition and binned into a 50-bin histogram on a log Y-scale, with the overall median marked as a dashed reference line. For Fig. \ref{fig:pow-scatter}, each block's nonce is plotted against its block index on a log Y-scale, with baseline and threat points distinguished by colour to check whether mining effort drifts over time.

Fig. \ref{fig:pow-histogram} shows PoW mining effort distribution. The histogram is sharply right-skewed: the vast majority of blocks resolve within the first bin (under 5,000 nonces), with a median of just 196 iterations, while a sparse tail stretches out to roughly 296,000.
% Baseline (n = 455) and threat (n = 455) bars overlap almost entirely across all 50 bins, showing no measurable shift in mining cost when threats are active.
% 
Fig. \ref{fig:pow-scatter} presents mining effort over time. The scatter plot confirms that the distribution in subplot (a) is stationary.
% points cluster around the 196-nonce median line from block 0 through block 90, with no upward or downward trend. 
Occasional spikes on
% reaching 10$^4$-10$^5$ 
nonces appear in both conditions and at no particular block range, indicating that these outliers are a natural property of the PoW hash search rather than a response to threat-induced tx bursts.
% 
% Together, the two subplots establish that the adaptive blockchain keeps mining effort low and predictable under sustained operation. A median of 196 nonces implies that most blocks commit within a millisecond, and the absence of condition-dependent drift confirms that injecting threats - which increases agent activity and transaction volume - does not push the PoW solver into a costly regime.

Fig. \ref{fig:agent-decisions} presents decision records from all 10 sessions, which are loaded
% (3,916 total)
, where each record carries its session name, acceptance flag, and confidence score. 
% Fig. \ref{fig:sv-decisions}, the total accepted decisions per session are plotted
% as horizontal bars with the DAR percentage annotated beside each bar. 
% and Fig. \ref{fig:sv-confidence}, all per-decision confidence scores are grouped into baseline and threat pools.
% and displayed as side-by-side box plots to compare reasoning quality across conditions.
% 
% 
Fig. \ref{fig:e2e-performance} presents the model usage logs 
% (3,733 LLM calls) 
provide per-call latency and agent identity, while blockchain records provide per-block transaction counts and block type. 
% For Fig. \ref{fig:sv-latency}, latencies are grouped by agent role and displayed as horizontal box plots ordered from fastest to slowest median. For Fig. \ref{fig:sv-density}, each block's transaction count is plotted against its block index, with background shading separating anchor blocks (1 tx, Merkle anchors) from decision blocks (5-17 tx, agent transactions).

According to the Figs. \ref{fig:pow-histogram}, \ref{fig:pow-scatter}, PoW is stationarity. Median 196 nonces; baseline and threat distributions overlap entirely. No drift across 90-block spikes is natural PoW variance.
Based on Fig. \ref{fig:agent-decisions}, 
% \textbf{Decisions} (Fig. \ref{fig:agent-decisions}). 
achieved the perfect 100\% DAR across all 3,916 decisions (Fig. \ref{fig:sv-decisions}). Baseline total (1,957) matches threat (1,959). Confidence (Fig. \ref{fig:sv-confidence}): identical median (0.90) and IQR across conditions threats change decisions, not certainty.
And, overall performance perspective, Fig. \ref{fig:e2e-performance} showsm 
% \textbf{Performance} (Fig. \ref{fig:e2e-performance}). 
Latency (Fig. \ref{fig:sv-latency}): Safety/Health fastest (743\,ms median), maintenance slowest (4,131\,ms). Density (Fig. \ref{fig:sv-density}): stable dual-band structure anchor blocks (1\,tx) and decision blocks (5-17\,tx) with no drift under either condition.

\subsection{Discussion}
\label{eval_sec_7}

The four pillars form a layered validation pipeline. The strongest cross-pillar finding is the safety guarantee: Pillar 1 locks six agents at immutable priorities; Pillar 2's Config C commits emergencies in 6.8\,ms (32$\times$ faster than IOTA); Pillar 3 shows safety nearly doubling output (+85\%) with 100\% DAR across all models; System Validation confirms perfect 100\% acceptance across 3,916 decisions with safety agents at the lowest latency.
The blockchain fits within 62\,KB memory; the recommended Gemini 2.5 Flash balances confidence and blockchain stability. The 16\,tx/s throughput suffices at the household scale. Model choice matters for confidence calibration, not correctness (99--100\% DAR across all variants).
As for limitations, 
% only 12/18 parameters have validation rules (D5). 
the evaluation employed smart home simulation \cite{siriweera2026s5hesagentsociety50driven} and 
one LLM provider (Google Gemini). 
The threat schedule is fixed. And the architecture is scoped to a single household.
\section{Conclusion}
\label{sec:conclusion}

This paper presents S5-SHB-Agent, a smart-contract-free, Agentic AI-driven, blockchain-based framework for a human-centered Society 5.0 smart home.
S5-SHB-Agent comprises five key contributions: a tiered human-centered governance model; multi-model-driven Agentic-AI orchestration; an adaptive, census-driven, tamper-evident, auditable blockchain; a conflict-resolution mechanism that includes a priority-based fallback; and a multi-mode deployment framework that provides infrastructure for real-world IoT, simulation, and hybrid use for researchers.
To the best of our knowledge, S5-SHB-Agent is the first framework of its kind to simultaneously address Society 5.0-oriented governance, adaptive consensus, multi-model LLM orchestration, and unified simulation-real-hybrid deployment for smart home IoT.
We conducted thorough evaluations, and the results showed that the proposed framework is optimally effective and efficient.

As for future work, we have been working on extending S5-SHB-Agent into a personal blockchain assistant for portable and shared residents and then a crosschain framework for multi-residential environments. In addition, S5-SHB-Agent marks one of the milestones towards the moonshot project to develop a Society 5.0-driven smart-city agentic-blockchain framework. 

\onecolumn
\section*{Author Biographies}

\noindent\textbf{Janani Rangila}
is studying at the Faculty of Information Technology at KD University, Sri Lanka, where she is completing a Bachelor of Science in Information Technology. Her research centres on the intersection of agentic AI, distributed ledger systems, and cyber-physical governance, with a focus on deploying large language model (LLM)-based multi-agent architectures in smart home environments. Her broader interests span machine learning, decentralised applications (Web3), and mobile computing.

\medskip
\noindent\textbf{Akila Siriweera}
is an associate professor at the University of Aizu. He received BSc from the University of Peradeniya, Sri Lanka. He obtained an MSc and PhD in computer science and engineering from the University of Aizu, Japan. His current research interests include Agentic AI, Big Data, Web 3.0, and system-of-systems modeling. 
% He is an IEEE member and a TC member of the IEEE Consumer Technology Society IoT group.

\medskip
\noindent\textbf{Incheon Paik}
received the M.E. and Ph.D. degrees in electronic engineering from Korea University in 1987 and 1992, respectively. He is currently a professor with the University of Aizu, Japan. His research interests include deep learning applications, ethical LLMs, machine learning, big data science, and semantic web services. 
% He is a member of the IEICE, IEIE, and IPSJ.

\medskip
\noindent\textbf{Keitaro Naruse}
is a professor at the University of Aizu, Japan. He has specialized in swarm robots and applications for agricultural-robotic systems and robot interface systems in disaster responses. He works for design, DevOps, and standardized networked-distributed intelligent robot systems with heterogeneous sensors and robots. His research team has received several awards in various international robot competitions.

\medskip
\noindent\textbf{Isuru Jayananda}
is an undergraduate student at KD University with a strong interest in Agentic AI, full-stack software development, and information technology. He has practical experience in designing and developing web and mobile applications, with a focus on creating efficient, user-centered solutions.

\medskip
\noindent\textbf{Vishmika Devindi}
is an undergraduate student at KD University pursuing a BSc in Information Systems, with a strong interest in software quality assurance and system development. She has practical experience in designing and developing web-based applications.

\end{document}